\def\year{2019}\relax
\documentclass[letterpaper]{article} 
\usepackage{aaai19}  
\usepackage{times}  
\usepackage{helvet} 
\usepackage{courier}  
\usepackage[hyphens]{url}  
\usepackage{graphicx} 
\usepackage{graphicx}  
\usepackage{url}
\usepackage{subcaption}
\usepackage{latexsym}
\usepackage{array}
\usepackage[font=small]{caption}
\usepackage{multirow}
\usepackage{comment}
\usepackage{amsfonts}
\usepackage{enumerate}
\usepackage{enumitem}
\usepackage{xcolor}
\usepackage{color}

\usepackage{graphicx,amsmath}
\usepackage[utf8]{inputenc}
\usepackage[export]{adjustbox}

\makeatletter

\def\private@AppleEmoji#1{%
	\def\private@AppleEmojibaseemoji{#1}
	\private@AppleEmojipae} 

\edef\private@AppleEmojimodifierbyteone{\noexpand\UTFviii@four@octets\string^^f0}
\edef\private@AppleEmojimoifierlastthreebytesb{\string^^9f\string^^8f\string^^bb}
\edef\private@AppleEmojimoifierlastthreebytesc{\string^^9f\string^^8f\string^^bc}
\edef\private@AppleEmojimoifierlastthreebytesd{\string^^9f\string^^8f\string^^bd}
\edef\private@AppleEmojimoifierlastthreebytese{\string^^9f\string^^8f\string^^be}
\edef\private@AppleEmojimoifierlastthreebytesf{\string^^9f\string^^8f\string^^bf}

\def\private@AppleEmojipae{\futurelet\private@AppleEmojitmp\private@AppleEmojipaex}

\def\private@AppleEmojipaex{%
	\ifx\private@AppleEmojitmp\private@AppleEmojimodifierbyteone 
	\expandafter\private@AppleEmojigetnextthreebytes 
	\else
	\private@AppleEmojiprint\private@AppleEmojibaseemoji{}
	\fi
}

\def\private@AppleEmojigetnextthreebytes#1#2#3#4{%
	\edef\private@AppleEmojitmpb{\string#2\string#3\string#4}
	\ifx\private@AppleEmojitmpb\private@AppleEmojimoifierlastthreebytesb
	\private@AppleEmojiprint\private@AppleEmojibaseemoji{1F3FB}%
	\else
	\ifx\private@AppleEmojitmpb\private@AppleEmojimoifierlastthreebytesc
	\private@AppleEmojiprint\private@AppleEmojibaseemoji{1F3FC}%
	\else
	\ifx\private@AppleEmojitmpb\private@AppleEmojimoifierlastthreebytesd
	\private@AppleEmojiprint\private@AppleEmojibaseemoji{1F3FD}%
	\else
	\ifx\private@AppleEmojitmpb\private@AppleEmojimoifierlastthreebytese
	\private@AppleEmojiprint\private@AppleEmojibaseemoji{1F3FE}%
	\else
	\ifx\private@AppleEmojitmpb\private@AppleEmojimoifierlastthreebytesf
	\private@AppleEmojiprint\private@AppleEmojibaseemoji{1F3FF}%
	\else
	\private@AppleEmojiprint\private@AppleEmojibaseemoji{}\private@AppleEmojimodifierbyteone#2#3#4%
	\fi
	\fi
	\fi
	\fi
	\fi}

\def\private@AppleEmojiprint#1#2
{%
	\text{%
		\includegraphics[height=1em,valign=B,raise=-0.2em]{#1#2.png}%
	}%
}

\DeclareUnicodeCharacter{1F600}{\private@AppleEmoji{1F600}}
\DeclareUnicodeCharacter{1F466}{\private@AppleEmoji{1F466}}
\DeclareUnicodeCharacter{1F601}{\private@AppleEmoji{1F601}}
\DeclareUnicodeCharacter{1F602}{\private@AppleEmoji{1F602}}
\DeclareUnicodeCharacter{1F603}{\private@AppleEmoji{1F603}}
\DeclareUnicodeCharacter{1F604}{\private@AppleEmoji{1F604}}
\DeclareUnicodeCharacter{1F605}{\private@AppleEmoji{1F605}}
\DeclareUnicodeCharacter{1F606}{\private@AppleEmoji{1F606}}
\DeclareUnicodeCharacter{1F609}{\private@AppleEmoji{1F609}}
\DeclareUnicodeCharacter{1F60A}{\private@AppleEmoji{1F60A}}
\DeclareUnicodeCharacter{1F60B}{\private@AppleEmoji{1F60B}}
\DeclareUnicodeCharacter{1F60E}{\private@AppleEmoji{1F60E}}
\DeclareUnicodeCharacter{1F60D}{\private@AppleEmoji{1F60D}}
\DeclareUnicodeCharacter{1F618}{\private@AppleEmoji{1F618}}
\DeclareUnicodeCharacter{1F617}{\private@AppleEmoji{1F617}}
\DeclareUnicodeCharacter{1F619}{\private@AppleEmoji{1F619}}
\DeclareUnicodeCharacter{1F61A}{\private@AppleEmoji{1F61A}}
\DeclareUnicodeCharacter{263A}{\private@AppleEmoji{263A}}
\DeclareUnicodeCharacter{1F642}{\private@AppleEmoji{1F642}}
\DeclareUnicodeCharacter{1F917}{\private@AppleEmoji{1F917}}
\DeclareUnicodeCharacter{1F607}{\private@AppleEmoji{1F607}}
\DeclareUnicodeCharacter{1F914}{\private@AppleEmoji{1F914}}
\DeclareUnicodeCharacter{1F610}{\private@AppleEmoji{1F610}}
\DeclareUnicodeCharacter{1F611}{\private@AppleEmoji{1F611}}
\DeclareUnicodeCharacter{1F636}{\private@AppleEmoji{1F636}}
\DeclareUnicodeCharacter{1F644}{\private@AppleEmoji{1F644}}
\DeclareUnicodeCharacter{1F60F}{\private@AppleEmoji{1F60F}}
\DeclareUnicodeCharacter{1F623}{\private@AppleEmoji{1F623}}
\DeclareUnicodeCharacter{1F625}{\private@AppleEmoji{1F625}}
\DeclareUnicodeCharacter{1F62E}{\private@AppleEmoji{1F62E}}
\DeclareUnicodeCharacter{1F910}{\private@AppleEmoji{1F910}}
\DeclareUnicodeCharacter{1F62F}{\private@AppleEmoji{1F62F}}
\DeclareUnicodeCharacter{1F62A}{\private@AppleEmoji{1F62A}}
\DeclareUnicodeCharacter{1F62B}{\private@AppleEmoji{1F62B}}
\DeclareUnicodeCharacter{1F634}{\private@AppleEmoji{1F634}}
\DeclareUnicodeCharacter{1F60C}{\private@AppleEmoji{1F60C}}
\DeclareUnicodeCharacter{1F913}{\private@AppleEmoji{1F913}}
\DeclareUnicodeCharacter{1F61B}{\private@AppleEmoji{1F61B}}
\DeclareUnicodeCharacter{1F61C}{\private@AppleEmoji{1F61C}}
\DeclareUnicodeCharacter{1F61D}{\private@AppleEmoji{1F61D}}
\DeclareUnicodeCharacter{2639}{\private@AppleEmoji{2639}}
\DeclareUnicodeCharacter{1F641}{\private@AppleEmoji{1F641}}
\DeclareUnicodeCharacter{1F612}{\private@AppleEmoji{1F612}}
\DeclareUnicodeCharacter{1F613}{\private@AppleEmoji{1F613}}
\DeclareUnicodeCharacter{1F614}{\private@AppleEmoji{1F614}}
\DeclareUnicodeCharacter{1F615}{\private@AppleEmoji{1F615}}
\DeclareUnicodeCharacter{1F616}{\private@AppleEmoji{1F616}}
\DeclareUnicodeCharacter{1F643}{\private@AppleEmoji{1F643}}
\DeclareUnicodeCharacter{1F637}{\private@AppleEmoji{1F637}}
\DeclareUnicodeCharacter{1F912}{\private@AppleEmoji{1F912}}
\DeclareUnicodeCharacter{1F915}{\private@AppleEmoji{1F915}}
\DeclareUnicodeCharacter{1F911}{\private@AppleEmoji{1F911}}
\DeclareUnicodeCharacter{1F632}{\private@AppleEmoji{1F632}}
\DeclareUnicodeCharacter{1F61E}{\private@AppleEmoji{1F61E}}
\DeclareUnicodeCharacter{1F61F}{\private@AppleEmoji{1F61F}}
\DeclareUnicodeCharacter{1F624}{\private@AppleEmoji{1F624}}
\DeclareUnicodeCharacter{1F622}{\private@AppleEmoji{1F622}}
\DeclareUnicodeCharacter{1F62D}{\private@AppleEmoji{1F62D}}
\DeclareUnicodeCharacter{1F626}{\private@AppleEmoji{1F626}}
\DeclareUnicodeCharacter{1F627}{\private@AppleEmoji{1F627}}
\DeclareUnicodeCharacter{1F628}{\private@AppleEmoji{1F628}}
\DeclareUnicodeCharacter{1F629}{\private@AppleEmoji{1F629}}
\DeclareUnicodeCharacter{1F62C}{\private@AppleEmoji{1F62C}}
\DeclareUnicodeCharacter{1F630}{\private@AppleEmoji{1F630}}
\DeclareUnicodeCharacter{1F631}{\private@AppleEmoji{1F631}}
\DeclareUnicodeCharacter{1F633}{\private@AppleEmoji{1F633}}
\DeclareUnicodeCharacter{1F635}{\private@AppleEmoji{1F635}}
\DeclareUnicodeCharacter{1F621}{\private@AppleEmoji{1F621}}
\DeclareUnicodeCharacter{1F620}{\private@AppleEmoji{1F620}}
\DeclareUnicodeCharacter{1F608}{\private@AppleEmoji{1F608}}
\DeclareUnicodeCharacter{1F47F}{\private@AppleEmoji{1F47F}}
\DeclareUnicodeCharacter{1F479}{\private@AppleEmoji{1F479}}
\DeclareUnicodeCharacter{1F47A}{\private@AppleEmoji{1F47A}}
\DeclareUnicodeCharacter{1F480}{\private@AppleEmoji{1F480}}
\DeclareUnicodeCharacter{2620}{\private@AppleEmoji{2620}}
\DeclareUnicodeCharacter{1F47B}{\private@AppleEmoji{1F47B}}
\DeclareUnicodeCharacter{1F47D}{\private@AppleEmoji{1F47D}}
\DeclareUnicodeCharacter{1F47E}{\private@AppleEmoji{1F47E}}
\DeclareUnicodeCharacter{1F916}{\private@AppleEmoji{1F916}}
\DeclareUnicodeCharacter{1F4A9}{\private@AppleEmoji{1F4A9}}
\DeclareUnicodeCharacter{1F63A}{\private@AppleEmoji{1F63A}}
\DeclareUnicodeCharacter{1F638}{\private@AppleEmoji{1F638}}
\DeclareUnicodeCharacter{1F639}{\private@AppleEmoji{1F639}}
\DeclareUnicodeCharacter{1F63B}{\private@AppleEmoji{1F63B}}
\DeclareUnicodeCharacter{1F63C}{\private@AppleEmoji{1F63C}}
\DeclareUnicodeCharacter{1F63D}{\private@AppleEmoji{1F63D}}
\DeclareUnicodeCharacter{1F640}{\private@AppleEmoji{1F640}}
\DeclareUnicodeCharacter{1F63F}{\private@AppleEmoji{1F63F}}
\DeclareUnicodeCharacter{1F63E}{\private@AppleEmoji{1F63E}}
\DeclareUnicodeCharacter{1F648}{\private@AppleEmoji{1F648}}
\DeclareUnicodeCharacter{1F649}{\private@AppleEmoji{1F649}}
\DeclareUnicodeCharacter{1F64A}{\private@AppleEmoji{1F64A}}
\DeclareUnicodeCharacter{1F467}{\private@AppleEmoji{1F467}}
\DeclareUnicodeCharacter{1F468}{\private@AppleEmoji{1F468}}
\DeclareUnicodeCharacter{1F469}{\private@AppleEmoji{1F469}}
\DeclareUnicodeCharacter{1F474}{\private@AppleEmoji{1F474}}
\DeclareUnicodeCharacter{1F475}{\private@AppleEmoji{1F475}}
\DeclareUnicodeCharacter{1F476}{\private@AppleEmoji{1F476}}
\DeclareUnicodeCharacter{1F471}{\private@AppleEmoji{1F471}}
\DeclareUnicodeCharacter{1F46E}{\private@AppleEmoji{1F46E}}
\DeclareUnicodeCharacter{1F472}{\private@AppleEmoji{1F472}}
\DeclareUnicodeCharacter{1F473}{\private@AppleEmoji{1F473}}
\DeclareUnicodeCharacter{1F477}{\private@AppleEmoji{1F477}}
\DeclareUnicodeCharacter{26D1}{\private@AppleEmoji{26D1}}
\DeclareUnicodeCharacter{1F478}{\private@AppleEmoji{1F478}}
\DeclareUnicodeCharacter{1F482}{\private@AppleEmoji{1F482}}
\DeclareUnicodeCharacter{1F575}{\private@AppleEmoji{1F575}}
\DeclareUnicodeCharacter{1F385}{\private@AppleEmoji{1F385}}
\DeclareUnicodeCharacter{1F47C}{\private@AppleEmoji{1F47C}}
\DeclareUnicodeCharacter{1F486}{\private@AppleEmoji{1F486}}
\DeclareUnicodeCharacter{1F487}{\private@AppleEmoji{1F487}}
\DeclareUnicodeCharacter{1F470}{\private@AppleEmoji{1F470}}
\DeclareUnicodeCharacter{1F64D}{\private@AppleEmoji{1F64D}}
\DeclareUnicodeCharacter{1F64E}{\private@AppleEmoji{1F64E}}
\DeclareUnicodeCharacter{1F645}{\private@AppleEmoji{1F645}}
\DeclareUnicodeCharacter{1F646}{\private@AppleEmoji{1F646}}
\DeclareUnicodeCharacter{1F481}{\private@AppleEmoji{1F481}}
\DeclareUnicodeCharacter{1F64B}{\private@AppleEmoji{1F64B}}
\DeclareUnicodeCharacter{1F647}{\private@AppleEmoji{1F647}}
\DeclareUnicodeCharacter{1F64C}{\private@AppleEmoji{1F64C}}
\DeclareUnicodeCharacter{1F64F}{\private@AppleEmoji{1F64F}}
\DeclareUnicodeCharacter{1F5E3}{\private@AppleEmoji{1F5E3}}
\DeclareUnicodeCharacter{1F464}{\private@AppleEmoji{1F464}}
\DeclareUnicodeCharacter{1F465}{\private@AppleEmoji{1F465}}
\DeclareUnicodeCharacter{1F6B6}{\private@AppleEmoji{1F6B6}}
\DeclareUnicodeCharacter{1F3C3}{\private@AppleEmoji{1F3C3}}
\DeclareUnicodeCharacter{1F46F}{\private@AppleEmoji{1F46F}}
\DeclareUnicodeCharacter{1F483}{\private@AppleEmoji{1F483}}
\DeclareUnicodeCharacter{1F574}{\private@AppleEmoji{1F574}}
\DeclareUnicodeCharacter{1F48F}{\private@AppleEmoji{1F48F}}
\DeclareUnicodeCharacter{1F491}{\private@AppleEmoji{1F491}}
\DeclareUnicodeCharacter{1F46A}{\private@AppleEmoji{1F46A}}
\DeclareUnicodeCharacter{1F46B}{\private@AppleEmoji{1F46B}}
\DeclareUnicodeCharacter{1F46C}{\private@AppleEmoji{1F46C}}
\DeclareUnicodeCharacter{1F46D}{\private@AppleEmoji{1F46D}}
\DeclareUnicodeCharacter{1F3FB}{\private@AppleEmoji{1F3FB}}
\DeclareUnicodeCharacter{1F3FC}{\private@AppleEmoji{1F3FC}}
\DeclareUnicodeCharacter{1F3FD}{\private@AppleEmoji{1F3FD}}
\DeclareUnicodeCharacter{1F3FE}{\private@AppleEmoji{1F3FE}}
\DeclareUnicodeCharacter{1F3FF}{\private@AppleEmoji{1F3FF}}
\DeclareUnicodeCharacter{1F4AA}{\private@AppleEmoji{1F4AA}}
\DeclareUnicodeCharacter{1F448}{\private@AppleEmoji{1F448}}
\DeclareUnicodeCharacter{1F449}{\private@AppleEmoji{1F449}}
\DeclareUnicodeCharacter{261D}{\private@AppleEmoji{261D}}
\DeclareUnicodeCharacter{1F446}{\private@AppleEmoji{1F446}}
\DeclareUnicodeCharacter{1F595}{\private@AppleEmoji{1F595}}
\DeclareUnicodeCharacter{1F447}{\private@AppleEmoji{1F447}}
\DeclareUnicodeCharacter{270C}{\private@AppleEmoji{270C}}
\DeclareUnicodeCharacter{1F596}{\private@AppleEmoji{1F596}}
\DeclareUnicodeCharacter{1F918}{\private@AppleEmoji{1F918}}
\DeclareUnicodeCharacter{1F590}{\private@AppleEmoji{1F590}}
\DeclareUnicodeCharacter{270B}{\private@AppleEmoji{270B}}
\DeclareUnicodeCharacter{1F44C}{\private@AppleEmoji{1F44C}}
\DeclareUnicodeCharacter{1F44D}{\private@AppleEmoji{1F44D}}
\DeclareUnicodeCharacter{1F44E}{\private@AppleEmoji{1F44E}}
\DeclareUnicodeCharacter{270A}{\private@AppleEmoji{270A}}
\DeclareUnicodeCharacter{1F44A}{\private@AppleEmoji{1F44A}}
\DeclareUnicodeCharacter{1F44B}{\private@AppleEmoji{1F44B}}
\DeclareUnicodeCharacter{1F44F}{\private@AppleEmoji{1F44F}}
\DeclareUnicodeCharacter{1F450}{\private@AppleEmoji{1F450}}
\DeclareUnicodeCharacter{270D}{\private@AppleEmoji{270D}}
\DeclareUnicodeCharacter{1F485}{\private@AppleEmoji{1F485}}
\DeclareUnicodeCharacter{1F442}{\private@AppleEmoji{1F442}}
\DeclareUnicodeCharacter{1F443}{\private@AppleEmoji{1F443}}
\DeclareUnicodeCharacter{1F463}{\private@AppleEmoji{1F463}}
\DeclareUnicodeCharacter{1F440}{\private@AppleEmoji{1F440}}
\DeclareUnicodeCharacter{1F441}{\private@AppleEmoji{1F441}}
\DeclareUnicodeCharacter{1F445}{\private@AppleEmoji{1F445}}
\DeclareUnicodeCharacter{1F444}{\private@AppleEmoji{1F444}}
\DeclareUnicodeCharacter{1F48B}{\private@AppleEmoji{1F48B}}
\DeclareUnicodeCharacter{1F498}{\private@AppleEmoji{1F498}}
\DeclareUnicodeCharacter{2764}{\private@AppleEmoji{2764}}
\DeclareUnicodeCharacter{1F493}{\private@AppleEmoji{1F493}}
\DeclareUnicodeCharacter{1F494}{\private@AppleEmoji{1F494}}
\DeclareUnicodeCharacter{1F495}{\private@AppleEmoji{1F495}}
\DeclareUnicodeCharacter{1F496}{\private@AppleEmoji{1F496}}
\DeclareUnicodeCharacter{1F497}{\private@AppleEmoji{1F497}}
\DeclareUnicodeCharacter{1F499}{\private@AppleEmoji{1F499}}
\DeclareUnicodeCharacter{1F49A}{\private@AppleEmoji{1F49A}}
\DeclareUnicodeCharacter{1F49B}{\private@AppleEmoji{1F49B}}
\DeclareUnicodeCharacter{1F49C}{\private@AppleEmoji{1F49C}}
\DeclareUnicodeCharacter{1F49D}{\private@AppleEmoji{1F49D}}
\DeclareUnicodeCharacter{1F49E}{\private@AppleEmoji{1F49E}}
\DeclareUnicodeCharacter{1F49F}{\private@AppleEmoji{1F49F}}
\DeclareUnicodeCharacter{2763}{\private@AppleEmoji{2763}}
\DeclareUnicodeCharacter{1F48C}{\private@AppleEmoji{1F48C}}
\DeclareUnicodeCharacter{1F4A4}{\private@AppleEmoji{1F4A4}}
\DeclareUnicodeCharacter{1F4A2}{\private@AppleEmoji{1F4A2}}
\DeclareUnicodeCharacter{1F4A3}{\private@AppleEmoji{1F4A3}}
\DeclareUnicodeCharacter{1F4A5}{\private@AppleEmoji{1F4A5}}
\DeclareUnicodeCharacter{1F4A6}{\private@AppleEmoji{1F4A6}}
\DeclareUnicodeCharacter{1F4A8}{\private@AppleEmoji{1F4A8}}
\DeclareUnicodeCharacter{1F4AB}{\private@AppleEmoji{1F4AB}}
\DeclareUnicodeCharacter{1F4AC}{\private@AppleEmoji{1F4AC}}
\DeclareUnicodeCharacter{1F5E8}{\private@AppleEmoji{1F5E8}}
\DeclareUnicodeCharacter{1F5EF}{\private@AppleEmoji{1F5EF}}
\DeclareUnicodeCharacter{1F4AD}{\private@AppleEmoji{1F4AD}}
\DeclareUnicodeCharacter{1F573}{\private@AppleEmoji{1F573}}
\DeclareUnicodeCharacter{1F453}{\private@AppleEmoji{1F453}}
\DeclareUnicodeCharacter{1F576}{\private@AppleEmoji{1F576}}
\DeclareUnicodeCharacter{1F454}{\private@AppleEmoji{1F454}}
\DeclareUnicodeCharacter{1F455}{\private@AppleEmoji{1F455}}
\DeclareUnicodeCharacter{1F456}{\private@AppleEmoji{1F456}}
\DeclareUnicodeCharacter{1F457}{\private@AppleEmoji{1F457}}
\DeclareUnicodeCharacter{1F458}{\private@AppleEmoji{1F458}}
\DeclareUnicodeCharacter{1F459}{\private@AppleEmoji{1F459}}
\DeclareUnicodeCharacter{1F45A}{\private@AppleEmoji{1F45A}}
\DeclareUnicodeCharacter{1F45B}{\private@AppleEmoji{1F45B}}
\DeclareUnicodeCharacter{1F45C}{\private@AppleEmoji{1F45C}}
\DeclareUnicodeCharacter{1F45D}{\private@AppleEmoji{1F45D}}
\DeclareUnicodeCharacter{1F6CD}{\private@AppleEmoji{1F6CD}}
\DeclareUnicodeCharacter{1F392}{\private@AppleEmoji{1F392}}
\DeclareUnicodeCharacter{1F45E}{\private@AppleEmoji{1F45E}}
\DeclareUnicodeCharacter{1F45F}{\private@AppleEmoji{1F45F}}
\DeclareUnicodeCharacter{1F460}{\private@AppleEmoji{1F460}}
\DeclareUnicodeCharacter{1F461}{\private@AppleEmoji{1F461}}
\DeclareUnicodeCharacter{1F462}{\private@AppleEmoji{1F462}}
\DeclareUnicodeCharacter{1F451}{\private@AppleEmoji{1F451}}
\DeclareUnicodeCharacter{1F452}{\private@AppleEmoji{1F452}}
\DeclareUnicodeCharacter{1F3A9}{\private@AppleEmoji{1F3A9}}
\DeclareUnicodeCharacter{1F393}{\private@AppleEmoji{1F393}}
\DeclareUnicodeCharacter{1F4FF}{\private@AppleEmoji{1F4FF}}
\DeclareUnicodeCharacter{1F484}{\private@AppleEmoji{1F484}}
\DeclareUnicodeCharacter{1F48D}{\private@AppleEmoji{1F48D}}
\DeclareUnicodeCharacter{1F48E}{\private@AppleEmoji{1F48E}}
\DeclareUnicodeCharacter{1F435}{\private@AppleEmoji{1F435}}
\DeclareUnicodeCharacter{1F412}{\private@AppleEmoji{1F412}}
\DeclareUnicodeCharacter{1F436}{\private@AppleEmoji{1F436}}
\DeclareUnicodeCharacter{1F415}{\private@AppleEmoji{1F415}}
\DeclareUnicodeCharacter{1F429}{\private@AppleEmoji{1F429}}
\DeclareUnicodeCharacter{1F43A}{\private@AppleEmoji{1F43A}}
\DeclareUnicodeCharacter{1F431}{\private@AppleEmoji{1F431}}
\DeclareUnicodeCharacter{1F408}{\private@AppleEmoji{1F408}}
\DeclareUnicodeCharacter{1F981}{\private@AppleEmoji{1F981}}
\DeclareUnicodeCharacter{1F42F}{\private@AppleEmoji{1F42F}}
\DeclareUnicodeCharacter{1F405}{\private@AppleEmoji{1F405}}
\DeclareUnicodeCharacter{1F406}{\private@AppleEmoji{1F406}}
\DeclareUnicodeCharacter{1F434}{\private@AppleEmoji{1F434}}
\DeclareUnicodeCharacter{1F40E}{\private@AppleEmoji{1F40E}}
\DeclareUnicodeCharacter{1F984}{\private@AppleEmoji{1F984}}
\DeclareUnicodeCharacter{1F42E}{\private@AppleEmoji{1F42E}}
\DeclareUnicodeCharacter{1F402}{\private@AppleEmoji{1F402}}
\DeclareUnicodeCharacter{1F403}{\private@AppleEmoji{1F403}}
\DeclareUnicodeCharacter{1F404}{\private@AppleEmoji{1F404}}
\DeclareUnicodeCharacter{1F437}{\private@AppleEmoji{1F437}}
\DeclareUnicodeCharacter{1F416}{\private@AppleEmoji{1F416}}
\DeclareUnicodeCharacter{1F417}{\private@AppleEmoji{1F417}}
\DeclareUnicodeCharacter{1F43D}{\private@AppleEmoji{1F43D}}
\DeclareUnicodeCharacter{1F40F}{\private@AppleEmoji{1F40F}}
\DeclareUnicodeCharacter{1F411}{\private@AppleEmoji{1F411}}
\DeclareUnicodeCharacter{1F410}{\private@AppleEmoji{1F410}}
\DeclareUnicodeCharacter{1F42A}{\private@AppleEmoji{1F42A}}
\DeclareUnicodeCharacter{1F42B}{\private@AppleEmoji{1F42B}}
\DeclareUnicodeCharacter{1F418}{\private@AppleEmoji{1F418}}
\DeclareUnicodeCharacter{1F42D}{\private@AppleEmoji{1F42D}}
\DeclareUnicodeCharacter{1F401}{\private@AppleEmoji{1F401}}
\DeclareUnicodeCharacter{1F400}{\private@AppleEmoji{1F400}}
\DeclareUnicodeCharacter{1F439}{\private@AppleEmoji{1F439}}
\DeclareUnicodeCharacter{1F430}{\private@AppleEmoji{1F430}}
\DeclareUnicodeCharacter{1F407}{\private@AppleEmoji{1F407}}
\DeclareUnicodeCharacter{1F43F}{\private@AppleEmoji{1F43F}}
\DeclareUnicodeCharacter{1F43B}{\private@AppleEmoji{1F43B}}
\DeclareUnicodeCharacter{1F428}{\private@AppleEmoji{1F428}}
\DeclareUnicodeCharacter{1F43C}{\private@AppleEmoji{1F43C}}
\DeclareUnicodeCharacter{1F43E}{\private@AppleEmoji{1F43E}}
\DeclareUnicodeCharacter{1F983}{\private@AppleEmoji{1F983}}
\DeclareUnicodeCharacter{1F414}{\private@AppleEmoji{1F414}}
\DeclareUnicodeCharacter{1F413}{\private@AppleEmoji{1F413}}
\DeclareUnicodeCharacter{1F423}{\private@AppleEmoji{1F423}}
\DeclareUnicodeCharacter{1F424}{\private@AppleEmoji{1F424}}
\DeclareUnicodeCharacter{1F425}{\private@AppleEmoji{1F425}}
\DeclareUnicodeCharacter{1F426}{\private@AppleEmoji{1F426}}
\DeclareUnicodeCharacter{1F427}{\private@AppleEmoji{1F427}}
\DeclareUnicodeCharacter{1F54A}{\private@AppleEmoji{1F54A}}
\DeclareUnicodeCharacter{1F438}{\private@AppleEmoji{1F438}}
\DeclareUnicodeCharacter{1F40A}{\private@AppleEmoji{1F40A}}
\DeclareUnicodeCharacter{1F422}{\private@AppleEmoji{1F422}}
\DeclareUnicodeCharacter{1F40D}{\private@AppleEmoji{1F40D}}
\DeclareUnicodeCharacter{1F432}{\private@AppleEmoji{1F432}}
\DeclareUnicodeCharacter{1F409}{\private@AppleEmoji{1F409}}
\DeclareUnicodeCharacter{1F433}{\private@AppleEmoji{1F433}}
\DeclareUnicodeCharacter{1F40B}{\private@AppleEmoji{1F40B}}
\DeclareUnicodeCharacter{1F42C}{\private@AppleEmoji{1F42C}}
\DeclareUnicodeCharacter{1F41F}{\private@AppleEmoji{1F41F}}
\DeclareUnicodeCharacter{1F420}{\private@AppleEmoji{1F420}}
\DeclareUnicodeCharacter{1F421}{\private@AppleEmoji{1F421}}
\DeclareUnicodeCharacter{1F419}{\private@AppleEmoji{1F419}}
\DeclareUnicodeCharacter{1F41A}{\private@AppleEmoji{1F41A}}
\DeclareUnicodeCharacter{1F980}{\private@AppleEmoji{1F980}}
\DeclareUnicodeCharacter{1F40C}{\private@AppleEmoji{1F40C}}
\DeclareUnicodeCharacter{1F41B}{\private@AppleEmoji{1F41B}}
\DeclareUnicodeCharacter{1F41C}{\private@AppleEmoji{1F41C}}
\DeclareUnicodeCharacter{1F41D}{\private@AppleEmoji{1F41D}}
\DeclareUnicodeCharacter{1F41E}{\private@AppleEmoji{1F41E}}
\DeclareUnicodeCharacter{1F577}{\private@AppleEmoji{1F577}}
\DeclareUnicodeCharacter{1F578}{\private@AppleEmoji{1F578}}
\DeclareUnicodeCharacter{1F982}{\private@AppleEmoji{1F982}}
\DeclareUnicodeCharacter{1F490}{\private@AppleEmoji{1F490}}
\DeclareUnicodeCharacter{1F338}{\private@AppleEmoji{1F338}}
\DeclareUnicodeCharacter{1F4AE}{\private@AppleEmoji{1F4AE}}
\DeclareUnicodeCharacter{1F3F5}{\private@AppleEmoji{1F3F5}}
\DeclareUnicodeCharacter{1F339}{\private@AppleEmoji{1F339}}
\DeclareUnicodeCharacter{1F33A}{\private@AppleEmoji{1F33A}}
\DeclareUnicodeCharacter{1F33B}{\private@AppleEmoji{1F33B}}
\DeclareUnicodeCharacter{1F33C}{\private@AppleEmoji{1F33C}}
\DeclareUnicodeCharacter{1F337}{\private@AppleEmoji{1F337}}
\DeclareUnicodeCharacter{1F331}{\private@AppleEmoji{1F331}}
\DeclareUnicodeCharacter{1F332}{\private@AppleEmoji{1F332}}
\DeclareUnicodeCharacter{1F333}{\private@AppleEmoji{1F333}}
\DeclareUnicodeCharacter{1F334}{\private@AppleEmoji{1F334}}
\DeclareUnicodeCharacter{1F335}{\private@AppleEmoji{1F335}}
\DeclareUnicodeCharacter{1F33E}{\private@AppleEmoji{1F33E}}
\DeclareUnicodeCharacter{1F33F}{\private@AppleEmoji{1F33F}}
\DeclareUnicodeCharacter{2618}{\private@AppleEmoji{2618}}
\DeclareUnicodeCharacter{1F340}{\private@AppleEmoji{1F340}}
\DeclareUnicodeCharacter{1F341}{\private@AppleEmoji{1F341}}
\DeclareUnicodeCharacter{1F342}{\private@AppleEmoji{1F342}}
\DeclareUnicodeCharacter{1F343}{\private@AppleEmoji{1F343}}
\DeclareUnicodeCharacter{1F347}{\private@AppleEmoji{1F347}}
\DeclareUnicodeCharacter{1F348}{\private@AppleEmoji{1F348}}
\DeclareUnicodeCharacter{1F349}{\private@AppleEmoji{1F349}}
\DeclareUnicodeCharacter{1F34A}{\private@AppleEmoji{1F34A}}
\DeclareUnicodeCharacter{1F34B}{\private@AppleEmoji{1F34B}}
\DeclareUnicodeCharacter{1F34C}{\private@AppleEmoji{1F34C}}
\DeclareUnicodeCharacter{1F34D}{\private@AppleEmoji{1F34D}}
\DeclareUnicodeCharacter{1F34E}{\private@AppleEmoji{1F34E}}
\DeclareUnicodeCharacter{1F34F}{\private@AppleEmoji{1F34F}}
\DeclareUnicodeCharacter{1F350}{\private@AppleEmoji{1F350}}
\DeclareUnicodeCharacter{1F351}{\private@AppleEmoji{1F351}}
\DeclareUnicodeCharacter{1F352}{\private@AppleEmoji{1F352}}
\DeclareUnicodeCharacter{1F353}{\private@AppleEmoji{1F353}}
\DeclareUnicodeCharacter{1F345}{\private@AppleEmoji{1F345}}
\DeclareUnicodeCharacter{1F346}{\private@AppleEmoji{1F346}}
\DeclareUnicodeCharacter{1F33D}{\private@AppleEmoji{1F33D}}
\DeclareUnicodeCharacter{1F336}{\private@AppleEmoji{1F336}}
\DeclareUnicodeCharacter{1F344}{\private@AppleEmoji{1F344}}
\DeclareUnicodeCharacter{1F330}{\private@AppleEmoji{1F330}}
\DeclareUnicodeCharacter{1F35E}{\private@AppleEmoji{1F35E}}
\DeclareUnicodeCharacter{1F9C0}{\private@AppleEmoji{1F9C0}}
\DeclareUnicodeCharacter{1F356}{\private@AppleEmoji{1F356}}
\DeclareUnicodeCharacter{1F357}{\private@AppleEmoji{1F357}}
\DeclareUnicodeCharacter{1F354}{\private@AppleEmoji{1F354}}
\DeclareUnicodeCharacter{1F35F}{\private@AppleEmoji{1F35F}}
\DeclareUnicodeCharacter{1F355}{\private@AppleEmoji{1F355}}
\DeclareUnicodeCharacter{1F32D}{\private@AppleEmoji{1F32D}}
\DeclareUnicodeCharacter{1F32E}{\private@AppleEmoji{1F32E}}
\DeclareUnicodeCharacter{1F32F}{\private@AppleEmoji{1F32F}}
\DeclareUnicodeCharacter{1F37F}{\private@AppleEmoji{1F37F}}
\DeclareUnicodeCharacter{1F372}{\private@AppleEmoji{1F372}}
\DeclareUnicodeCharacter{1F371}{\private@AppleEmoji{1F371}}
\DeclareUnicodeCharacter{1F358}{\private@AppleEmoji{1F358}}
\DeclareUnicodeCharacter{1F359}{\private@AppleEmoji{1F359}}
\DeclareUnicodeCharacter{1F35A}{\private@AppleEmoji{1F35A}}
\DeclareUnicodeCharacter{1F35B}{\private@AppleEmoji{1F35B}}
\DeclareUnicodeCharacter{1F35C}{\private@AppleEmoji{1F35C}}
\DeclareUnicodeCharacter{1F35D}{\private@AppleEmoji{1F35D}}
\DeclareUnicodeCharacter{1F360}{\private@AppleEmoji{1F360}}
\DeclareUnicodeCharacter{1F362}{\private@AppleEmoji{1F362}}
\DeclareUnicodeCharacter{1F363}{\private@AppleEmoji{1F363}}
\DeclareUnicodeCharacter{1F364}{\private@AppleEmoji{1F364}}
\DeclareUnicodeCharacter{1F365}{\private@AppleEmoji{1F365}}
\DeclareUnicodeCharacter{1F361}{\private@AppleEmoji{1F361}}
\DeclareUnicodeCharacter{1F366}{\private@AppleEmoji{1F366}}
\DeclareUnicodeCharacter{1F367}{\private@AppleEmoji{1F367}}
\DeclareUnicodeCharacter{1F368}{\private@AppleEmoji{1F368}}
\DeclareUnicodeCharacter{1F369}{\private@AppleEmoji{1F369}}
\DeclareUnicodeCharacter{1F36A}{\private@AppleEmoji{1F36A}}
\DeclareUnicodeCharacter{1F382}{\private@AppleEmoji{1F382}}
\DeclareUnicodeCharacter{1F370}{\private@AppleEmoji{1F370}}
\DeclareUnicodeCharacter{1F36B}{\private@AppleEmoji{1F36B}}
\DeclareUnicodeCharacter{1F36C}{\private@AppleEmoji{1F36C}}
\DeclareUnicodeCharacter{1F36D}{\private@AppleEmoji{1F36D}}
\DeclareUnicodeCharacter{1F36E}{\private@AppleEmoji{1F36E}}
\DeclareUnicodeCharacter{1F36F}{\private@AppleEmoji{1F36F}}
\DeclareUnicodeCharacter{1F37C}{\private@AppleEmoji{1F37C}}
\DeclareUnicodeCharacter{2615}{\private@AppleEmoji{2615}}
\DeclareUnicodeCharacter{1F375}{\private@AppleEmoji{1F375}}
\DeclareUnicodeCharacter{1F376}{\private@AppleEmoji{1F376}}
\DeclareUnicodeCharacter{1F37E}{\private@AppleEmoji{1F37E}}
\DeclareUnicodeCharacter{1F377}{\private@AppleEmoji{1F377}}
\DeclareUnicodeCharacter{1F378}{\private@AppleEmoji{1F378}}
\DeclareUnicodeCharacter{1F379}{\private@AppleEmoji{1F379}}
\DeclareUnicodeCharacter{1F37A}{\private@AppleEmoji{1F37A}}
\DeclareUnicodeCharacter{1F37B}{\private@AppleEmoji{1F37B}}
\DeclareUnicodeCharacter{1F37D}{\private@AppleEmoji{1F37D}}
\DeclareUnicodeCharacter{1F374}{\private@AppleEmoji{1F374}}
\DeclareUnicodeCharacter{1F373}{\private@AppleEmoji{1F373}}
\DeclareUnicodeCharacter{1F3FA}{\private@AppleEmoji{1F3FA}}
\DeclareUnicodeCharacter{1F30D}{\private@AppleEmoji{1F30D}}
\DeclareUnicodeCharacter{1F30E}{\private@AppleEmoji{1F30E}}
\DeclareUnicodeCharacter{1F30F}{\private@AppleEmoji{1F30F}}
\DeclareUnicodeCharacter{1F310}{\private@AppleEmoji{1F310}}
\DeclareUnicodeCharacter{1F5FA}{\private@AppleEmoji{1F5FA}}
\DeclareUnicodeCharacter{1F3D4}{\private@AppleEmoji{1F3D4}}
\DeclareUnicodeCharacter{26F0}{\private@AppleEmoji{26F0}}
\DeclareUnicodeCharacter{1F30B}{\private@AppleEmoji{1F30B}}
\DeclareUnicodeCharacter{1F5FB}{\private@AppleEmoji{1F5FB}}
\DeclareUnicodeCharacter{1F3D5}{\private@AppleEmoji{1F3D5}}
\DeclareUnicodeCharacter{1F3D6}{\private@AppleEmoji{1F3D6}}
\DeclareUnicodeCharacter{1F3DC}{\private@AppleEmoji{1F3DC}}
\DeclareUnicodeCharacter{1F3DD}{\private@AppleEmoji{1F3DD}}
\DeclareUnicodeCharacter{1F3DE}{\private@AppleEmoji{1F3DE}}
\DeclareUnicodeCharacter{1F3DF}{\private@AppleEmoji{1F3DF}}
\DeclareUnicodeCharacter{1F3DB}{\private@AppleEmoji{1F3DB}}
\DeclareUnicodeCharacter{1F3D7}{\private@AppleEmoji{1F3D7}}
\DeclareUnicodeCharacter{1F3D8}{\private@AppleEmoji{1F3D8}}
\DeclareUnicodeCharacter{1F3D9}{\private@AppleEmoji{1F3D9}}
\DeclareUnicodeCharacter{1F3DA}{\private@AppleEmoji{1F3DA}}
\DeclareUnicodeCharacter{1F3E0}{\private@AppleEmoji{1F3E0}}
\DeclareUnicodeCharacter{1F3E1}{\private@AppleEmoji{1F3E1}}
\DeclareUnicodeCharacter{26EA}{\private@AppleEmoji{26EA}}
\DeclareUnicodeCharacter{1F54B}{\private@AppleEmoji{1F54B}}
\DeclareUnicodeCharacter{1F54C}{\private@AppleEmoji{1F54C}}
\DeclareUnicodeCharacter{1F54D}{\private@AppleEmoji{1F54D}}
\DeclareUnicodeCharacter{26E9}{\private@AppleEmoji{26E9}}
\DeclareUnicodeCharacter{1F3E2}{\private@AppleEmoji{1F3E2}}
\DeclareUnicodeCharacter{1F3E3}{\private@AppleEmoji{1F3E3}}
\DeclareUnicodeCharacter{1F3E4}{\private@AppleEmoji{1F3E4}}
\DeclareUnicodeCharacter{1F3E5}{\private@AppleEmoji{1F3E5}}
\DeclareUnicodeCharacter{1F3E6}{\private@AppleEmoji{1F3E6}}
\DeclareUnicodeCharacter{1F3E8}{\private@AppleEmoji{1F3E8}}
\DeclareUnicodeCharacter{1F3E9}{\private@AppleEmoji{1F3E9}}
\DeclareUnicodeCharacter{1F3EA}{\private@AppleEmoji{1F3EA}}
\DeclareUnicodeCharacter{1F3EB}{\private@AppleEmoji{1F3EB}}
\DeclareUnicodeCharacter{1F3EC}{\private@AppleEmoji{1F3EC}}
\DeclareUnicodeCharacter{1F3ED}{\private@AppleEmoji{1F3ED}}
\DeclareUnicodeCharacter{1F3EF}{\private@AppleEmoji{1F3EF}}
\DeclareUnicodeCharacter{1F3F0}{\private@AppleEmoji{1F3F0}}
\DeclareUnicodeCharacter{1F492}{\private@AppleEmoji{1F492}}
\DeclareUnicodeCharacter{1F5FC}{\private@AppleEmoji{1F5FC}}
\DeclareUnicodeCharacter{1F5FD}{\private@AppleEmoji{1F5FD}}
\DeclareUnicodeCharacter{1F5FE}{\private@AppleEmoji{1F5FE}}
\DeclareUnicodeCharacter{26F2}{\private@AppleEmoji{26F2}}
\DeclareUnicodeCharacter{26FA}{\private@AppleEmoji{26FA}}
\DeclareUnicodeCharacter{1F301}{\private@AppleEmoji{1F301}}
\DeclareUnicodeCharacter{1F303}{\private@AppleEmoji{1F303}}
\DeclareUnicodeCharacter{1F304}{\private@AppleEmoji{1F304}}
\DeclareUnicodeCharacter{1F305}{\private@AppleEmoji{1F305}}
\DeclareUnicodeCharacter{1F306}{\private@AppleEmoji{1F306}}
\DeclareUnicodeCharacter{1F307}{\private@AppleEmoji{1F307}}
\DeclareUnicodeCharacter{1F309}{\private@AppleEmoji{1F309}}
\DeclareUnicodeCharacter{2668}{\private@AppleEmoji{2668}}
\DeclareUnicodeCharacter{1F30C}{\private@AppleEmoji{1F30C}}
\DeclareUnicodeCharacter{1F3A0}{\private@AppleEmoji{1F3A0}}
\DeclareUnicodeCharacter{1F3A1}{\private@AppleEmoji{1F3A1}}
\DeclareUnicodeCharacter{1F3A2}{\private@AppleEmoji{1F3A2}}
\DeclareUnicodeCharacter{1F488}{\private@AppleEmoji{1F488}}
\DeclareUnicodeCharacter{1F3AA}{\private@AppleEmoji{1F3AA}}
\DeclareUnicodeCharacter{1F3AD}{\private@AppleEmoji{1F3AD}}
\DeclareUnicodeCharacter{1F5BC}{\private@AppleEmoji{1F5BC}}
\DeclareUnicodeCharacter{1F3A8}{\private@AppleEmoji{1F3A8}}
\DeclareUnicodeCharacter{1F3B0}{\private@AppleEmoji{1F3B0}}
\DeclareUnicodeCharacter{1F682}{\private@AppleEmoji{1F682}}
\DeclareUnicodeCharacter{1F683}{\private@AppleEmoji{1F683}}
\DeclareUnicodeCharacter{1F684}{\private@AppleEmoji{1F684}}
\DeclareUnicodeCharacter{1F685}{\private@AppleEmoji{1F685}}
\DeclareUnicodeCharacter{1F686}{\private@AppleEmoji{1F686}}
\DeclareUnicodeCharacter{1F687}{\private@AppleEmoji{1F687}}
\DeclareUnicodeCharacter{1F688}{\private@AppleEmoji{1F688}}
\DeclareUnicodeCharacter{1F689}{\private@AppleEmoji{1F689}}
\DeclareUnicodeCharacter{1F68A}{\private@AppleEmoji{1F68A}}
\DeclareUnicodeCharacter{1F69D}{\private@AppleEmoji{1F69D}}
\DeclareUnicodeCharacter{1F69E}{\private@AppleEmoji{1F69E}}
\DeclareUnicodeCharacter{1F68B}{\private@AppleEmoji{1F68B}}
\DeclareUnicodeCharacter{1F68C}{\private@AppleEmoji{1F68C}}
\DeclareUnicodeCharacter{1F68D}{\private@AppleEmoji{1F68D}}
\DeclareUnicodeCharacter{1F68E}{\private@AppleEmoji{1F68E}}
\DeclareUnicodeCharacter{1F68F}{\private@AppleEmoji{1F68F}}
\DeclareUnicodeCharacter{1F690}{\private@AppleEmoji{1F690}}
\DeclareUnicodeCharacter{1F691}{\private@AppleEmoji{1F691}}
\DeclareUnicodeCharacter{1F692}{\private@AppleEmoji{1F692}}
\DeclareUnicodeCharacter{1F693}{\private@AppleEmoji{1F693}}
\DeclareUnicodeCharacter{1F694}{\private@AppleEmoji{1F694}}
\DeclareUnicodeCharacter{1F695}{\private@AppleEmoji{1F695}}
\DeclareUnicodeCharacter{1F696}{\private@AppleEmoji{1F696}}
\DeclareUnicodeCharacter{1F697}{\private@AppleEmoji{1F697}}
\DeclareUnicodeCharacter{1F698}{\private@AppleEmoji{1F698}}
\DeclareUnicodeCharacter{1F699}{\private@AppleEmoji{1F699}}
\DeclareUnicodeCharacter{1F69A}{\private@AppleEmoji{1F69A}}
\DeclareUnicodeCharacter{1F69B}{\private@AppleEmoji{1F69B}}
\DeclareUnicodeCharacter{1F69C}{\private@AppleEmoji{1F69C}}
\DeclareUnicodeCharacter{1F6B2}{\private@AppleEmoji{1F6B2}}
\DeclareUnicodeCharacter{26FD}{\private@AppleEmoji{26FD}}
\DeclareUnicodeCharacter{1F6E3}{\private@AppleEmoji{1F6E3}}
\DeclareUnicodeCharacter{1F6E4}{\private@AppleEmoji{1F6E4}}
\DeclareUnicodeCharacter{1F6A8}{\private@AppleEmoji{1F6A8}}
\DeclareUnicodeCharacter{1F6A5}{\private@AppleEmoji{1F6A5}}
\DeclareUnicodeCharacter{1F6A6}{\private@AppleEmoji{1F6A6}}
\DeclareUnicodeCharacter{1F6A7}{\private@AppleEmoji{1F6A7}}
\DeclareUnicodeCharacter{2693}{\private@AppleEmoji{2693}}
\DeclareUnicodeCharacter{26F5}{\private@AppleEmoji{26F5}}
\DeclareUnicodeCharacter{1F6A3}{\private@AppleEmoji{1F6A3}}
\DeclareUnicodeCharacter{1F6A4}{\private@AppleEmoji{1F6A4}}
\DeclareUnicodeCharacter{1F6F3}{\private@AppleEmoji{1F6F3}}
\DeclareUnicodeCharacter{26F4}{\private@AppleEmoji{26F4}}
\DeclareUnicodeCharacter{1F6E5}{\private@AppleEmoji{1F6E5}}
\DeclareUnicodeCharacter{1F6A2}{\private@AppleEmoji{1F6A2}}
\DeclareUnicodeCharacter{2708}{\private@AppleEmoji{2708}}
\DeclareUnicodeCharacter{1F6E9}{\private@AppleEmoji{1F6E9}}
\DeclareUnicodeCharacter{1F6EB}{\private@AppleEmoji{1F6EB}}
\DeclareUnicodeCharacter{1F6EC}{\private@AppleEmoji{1F6EC}}
\DeclareUnicodeCharacter{1F4BA}{\private@AppleEmoji{1F4BA}}
\DeclareUnicodeCharacter{1F681}{\private@AppleEmoji{1F681}}
\DeclareUnicodeCharacter{1F69F}{\private@AppleEmoji{1F69F}}
\DeclareUnicodeCharacter{1F6A0}{\private@AppleEmoji{1F6A0}}
\DeclareUnicodeCharacter{1F6A1}{\private@AppleEmoji{1F6A1}}
\DeclareUnicodeCharacter{1F680}{\private@AppleEmoji{1F680}}
\DeclareUnicodeCharacter{1F6F0}{\private@AppleEmoji{1F6F0}}
\DeclareUnicodeCharacter{1F6CE}{\private@AppleEmoji{1F6CE}}
\DeclareUnicodeCharacter{1F6AA}{\private@AppleEmoji{1F6AA}}
\DeclareUnicodeCharacter{1F6CC}{\private@AppleEmoji{1F6CC}}
\DeclareUnicodeCharacter{1F6CF}{\private@AppleEmoji{1F6CF}}
\DeclareUnicodeCharacter{1F6CB}{\private@AppleEmoji{1F6CB}}
\DeclareUnicodeCharacter{1F6BD}{\private@AppleEmoji{1F6BD}}
\DeclareUnicodeCharacter{1F6BF}{\private@AppleEmoji{1F6BF}}
\DeclareUnicodeCharacter{1F6C0}{\private@AppleEmoji{1F6C0}}
\DeclareUnicodeCharacter{1F6C1}{\private@AppleEmoji{1F6C1}}
\DeclareUnicodeCharacter{231B}{\private@AppleEmoji{231B}}
\DeclareUnicodeCharacter{23F3}{\private@AppleEmoji{23F3}}
\DeclareUnicodeCharacter{231A}{\private@AppleEmoji{231A}}
\DeclareUnicodeCharacter{23F0}{\private@AppleEmoji{23F0}}
\DeclareUnicodeCharacter{23F1}{\private@AppleEmoji{23F1}}
\DeclareUnicodeCharacter{23F2}{\private@AppleEmoji{23F2}}
\DeclareUnicodeCharacter{1F570}{\private@AppleEmoji{1F570}}
\DeclareUnicodeCharacter{1F55B}{\private@AppleEmoji{1F55B}}
\DeclareUnicodeCharacter{1F567}{\private@AppleEmoji{1F567}}
\DeclareUnicodeCharacter{1F550}{\private@AppleEmoji{1F550}}
\DeclareUnicodeCharacter{1F55C}{\private@AppleEmoji{1F55C}}
\DeclareUnicodeCharacter{1F551}{\private@AppleEmoji{1F551}}
\DeclareUnicodeCharacter{1F55D}{\private@AppleEmoji{1F55D}}
\DeclareUnicodeCharacter{1F552}{\private@AppleEmoji{1F552}}
\DeclareUnicodeCharacter{1F55E}{\private@AppleEmoji{1F55E}}
\DeclareUnicodeCharacter{1F553}{\private@AppleEmoji{1F553}}
\DeclareUnicodeCharacter{1F55F}{\private@AppleEmoji{1F55F}}
\DeclareUnicodeCharacter{1F554}{\private@AppleEmoji{1F554}}
\DeclareUnicodeCharacter{1F560}{\private@AppleEmoji{1F560}}
\DeclareUnicodeCharacter{1F555}{\private@AppleEmoji{1F555}}
\DeclareUnicodeCharacter{1F561}{\private@AppleEmoji{1F561}}
\DeclareUnicodeCharacter{1F556}{\private@AppleEmoji{1F556}}
\DeclareUnicodeCharacter{1F562}{\private@AppleEmoji{1F562}}
\DeclareUnicodeCharacter{1F557}{\private@AppleEmoji{1F557}}
\DeclareUnicodeCharacter{1F563}{\private@AppleEmoji{1F563}}
\DeclareUnicodeCharacter{1F558}{\private@AppleEmoji{1F558}}
\DeclareUnicodeCharacter{1F564}{\private@AppleEmoji{1F564}}
\DeclareUnicodeCharacter{1F559}{\private@AppleEmoji{1F559}}
\DeclareUnicodeCharacter{1F565}{\private@AppleEmoji{1F565}}
\DeclareUnicodeCharacter{1F55A}{\private@AppleEmoji{1F55A}}
\DeclareUnicodeCharacter{1F566}{\private@AppleEmoji{1F566}}
\DeclareUnicodeCharacter{1F311}{\private@AppleEmoji{1F311}}
\DeclareUnicodeCharacter{1F312}{\private@AppleEmoji{1F312}}
\DeclareUnicodeCharacter{1F313}{\private@AppleEmoji{1F313}}
\DeclareUnicodeCharacter{1F314}{\private@AppleEmoji{1F314}}
\DeclareUnicodeCharacter{1F315}{\private@AppleEmoji{1F315}}
\DeclareUnicodeCharacter{1F316}{\private@AppleEmoji{1F316}}
\DeclareUnicodeCharacter{1F317}{\private@AppleEmoji{1F317}}
\DeclareUnicodeCharacter{1F318}{\private@AppleEmoji{1F318}}
\DeclareUnicodeCharacter{1F319}{\private@AppleEmoji{1F319}}
\DeclareUnicodeCharacter{1F31A}{\private@AppleEmoji{1F31A}}
\DeclareUnicodeCharacter{1F31B}{\private@AppleEmoji{1F31B}}
\DeclareUnicodeCharacter{1F31C}{\private@AppleEmoji{1F31C}}
\DeclareUnicodeCharacter{1F321}{\private@AppleEmoji{1F321}}
\DeclareUnicodeCharacter{2600}{\private@AppleEmoji{2600}}
\DeclareUnicodeCharacter{1F31D}{\private@AppleEmoji{1F31D}}
\DeclareUnicodeCharacter{1F31E}{\private@AppleEmoji{1F31E}}
\DeclareUnicodeCharacter{2B50}{\private@AppleEmoji{2B50}}
\DeclareUnicodeCharacter{1F31F}{\private@AppleEmoji{1F31F}}
\DeclareUnicodeCharacter{1F320}{\private@AppleEmoji{1F320}}
\DeclareUnicodeCharacter{2601}{\private@AppleEmoji{2601}}
\DeclareUnicodeCharacter{26C5}{\private@AppleEmoji{26C5}}
\DeclareUnicodeCharacter{26C8}{\private@AppleEmoji{26C8}}
\DeclareUnicodeCharacter{1F324}{\private@AppleEmoji{1F324}}
\DeclareUnicodeCharacter{1F325}{\private@AppleEmoji{1F325}}
\DeclareUnicodeCharacter{1F326}{\private@AppleEmoji{1F326}}
\DeclareUnicodeCharacter{1F327}{\private@AppleEmoji{1F327}}
\DeclareUnicodeCharacter{1F328}{\private@AppleEmoji{1F328}}
\DeclareUnicodeCharacter{1F329}{\private@AppleEmoji{1F329}}
\DeclareUnicodeCharacter{1F32A}{\private@AppleEmoji{1F32A}}
\DeclareUnicodeCharacter{1F32B}{\private@AppleEmoji{1F32B}}
\DeclareUnicodeCharacter{1F32C}{\private@AppleEmoji{1F32C}}
\DeclareUnicodeCharacter{1F300}{\private@AppleEmoji{1F300}}
\DeclareUnicodeCharacter{1F308}{\private@AppleEmoji{1F308}}
\DeclareUnicodeCharacter{1F302}{\private@AppleEmoji{1F302}}
\DeclareUnicodeCharacter{2602}{\private@AppleEmoji{2602}}
\DeclareUnicodeCharacter{2614}{\private@AppleEmoji{2614}}
\DeclareUnicodeCharacter{26F1}{\private@AppleEmoji{26F1}}
\DeclareUnicodeCharacter{26A1}{\private@AppleEmoji{26A1}}
\DeclareUnicodeCharacter{2744}{\private@AppleEmoji{2744}}
\DeclareUnicodeCharacter{2603}{\private@AppleEmoji{2603}}
\DeclareUnicodeCharacter{26C4}{\private@AppleEmoji{26C4}}
\DeclareUnicodeCharacter{2604}{\private@AppleEmoji{2604}}
\DeclareUnicodeCharacter{1F525}{\private@AppleEmoji{1F525}}
\DeclareUnicodeCharacter{1F4A7}{\private@AppleEmoji{1F4A7}}
\DeclareUnicodeCharacter{1F30A}{\private@AppleEmoji{1F30A}}
\DeclareUnicodeCharacter{1F383}{\private@AppleEmoji{1F383}}
\DeclareUnicodeCharacter{1F384}{\private@AppleEmoji{1F384}}
\DeclareUnicodeCharacter{1F386}{\private@AppleEmoji{1F386}}
\DeclareUnicodeCharacter{1F387}{\private@AppleEmoji{1F387}}
\DeclareUnicodeCharacter{2728}{\private@AppleEmoji{2728}}
\DeclareUnicodeCharacter{1F388}{\private@AppleEmoji{1F388}}
\DeclareUnicodeCharacter{1F389}{\private@AppleEmoji{1F389}}
\DeclareUnicodeCharacter{1F38A}{\private@AppleEmoji{1F38A}}
\DeclareUnicodeCharacter{1F38B}{\private@AppleEmoji{1F38B}}
\DeclareUnicodeCharacter{1F38C}{\private@AppleEmoji{1F38C}}
\DeclareUnicodeCharacter{1F38D}{\private@AppleEmoji{1F38D}}
\DeclareUnicodeCharacter{1F38E}{\private@AppleEmoji{1F38E}}
\DeclareUnicodeCharacter{1F38F}{\private@AppleEmoji{1F38F}}
\DeclareUnicodeCharacter{1F390}{\private@AppleEmoji{1F390}}
\DeclareUnicodeCharacter{1F391}{\private@AppleEmoji{1F391}}
\DeclareUnicodeCharacter{1F380}{\private@AppleEmoji{1F380}}
\DeclareUnicodeCharacter{1F381}{\private@AppleEmoji{1F381}}
\DeclareUnicodeCharacter{1F396}{\private@AppleEmoji{1F396}}
\DeclareUnicodeCharacter{1F397}{\private@AppleEmoji{1F397}}
\DeclareUnicodeCharacter{1F39E}{\private@AppleEmoji{1F39E}}
\DeclareUnicodeCharacter{1F39F}{\private@AppleEmoji{1F39F}}
\DeclareUnicodeCharacter{1F3AB}{\private@AppleEmoji{1F3AB}}
\DeclareUnicodeCharacter{1F3F7}{\private@AppleEmoji{1F3F7}}
\DeclareUnicodeCharacter{26BD}{\private@AppleEmoji{26BD}}
\DeclareUnicodeCharacter{26BE}{\private@AppleEmoji{26BE}}
\DeclareUnicodeCharacter{1F3C0}{\private@AppleEmoji{1F3C0}}
\DeclareUnicodeCharacter{1F3C8}{\private@AppleEmoji{1F3C8}}
\DeclareUnicodeCharacter{1F3C9}{\private@AppleEmoji{1F3C9}}
\DeclareUnicodeCharacter{1F3BE}{\private@AppleEmoji{1F3BE}}
\DeclareUnicodeCharacter{1F3B1}{\private@AppleEmoji{1F3B1}}
\DeclareUnicodeCharacter{1F3B3}{\private@AppleEmoji{1F3B3}}
\DeclareUnicodeCharacter{26F3}{\private@AppleEmoji{26F3}}
\DeclareUnicodeCharacter{1F3CC}{\private@AppleEmoji{1F3CC}}
\DeclareUnicodeCharacter{26F8}{\private@AppleEmoji{26F8}}
\DeclareUnicodeCharacter{1F3A3}{\private@AppleEmoji{1F3A3}}
\DeclareUnicodeCharacter{1F3BD}{\private@AppleEmoji{1F3BD}}
\DeclareUnicodeCharacter{1F3BF}{\private@AppleEmoji{1F3BF}}
\DeclareUnicodeCharacter{26F7}{\private@AppleEmoji{26F7}}
\DeclareUnicodeCharacter{1F3C2}{\private@AppleEmoji{1F3C2}}
\DeclareUnicodeCharacter{1F3C4}{\private@AppleEmoji{1F3C4}}
\DeclareUnicodeCharacter{1F3C7}{\private@AppleEmoji{1F3C7}}
\DeclareUnicodeCharacter{1F3CA}{\private@AppleEmoji{1F3CA}}
\DeclareUnicodeCharacter{26F9}{\private@AppleEmoji{26F9}}
\DeclareUnicodeCharacter{1F3CB}{\private@AppleEmoji{1F3CB}}
\DeclareUnicodeCharacter{1F6B4}{\private@AppleEmoji{1F6B4}}
\DeclareUnicodeCharacter{1F6B5}{\private@AppleEmoji{1F6B5}}
\DeclareUnicodeCharacter{1F3CE}{\private@AppleEmoji{1F3CE}}
\DeclareUnicodeCharacter{1F3CD}{\private@AppleEmoji{1F3CD}}
\DeclareUnicodeCharacter{1F3C5}{\private@AppleEmoji{1F3C5}}
\DeclareUnicodeCharacter{1F3C6}{\private@AppleEmoji{1F3C6}}
\DeclareUnicodeCharacter{1F3CF}{\private@AppleEmoji{1F3CF}}
\DeclareUnicodeCharacter{1F3D0}{\private@AppleEmoji{1F3D0}}
\DeclareUnicodeCharacter{1F3D1}{\private@AppleEmoji{1F3D1}}
\DeclareUnicodeCharacter{1F3D2}{\private@AppleEmoji{1F3D2}}
\DeclareUnicodeCharacter{1F3D3}{\private@AppleEmoji{1F3D3}}
\DeclareUnicodeCharacter{1F3F8}{\private@AppleEmoji{1F3F8}}
\DeclareUnicodeCharacter{1F3AF}{\private@AppleEmoji{1F3AF}}
\DeclareUnicodeCharacter{1F3AE}{\private@AppleEmoji{1F3AE}}
\DeclareUnicodeCharacter{1F579}{\private@AppleEmoji{1F579}}
\DeclareUnicodeCharacter{1F3B2}{\private@AppleEmoji{1F3B2}}
\DeclareUnicodeCharacter{2660}{\private@AppleEmoji{2660}}
\DeclareUnicodeCharacter{2665}{\private@AppleEmoji{2665}}
\DeclareUnicodeCharacter{2666}{\private@AppleEmoji{2666}}
\DeclareUnicodeCharacter{2663}{\private@AppleEmoji{2663}}
\DeclareUnicodeCharacter{1F0CF}{\private@AppleEmoji{1F0CF}}
\DeclareUnicodeCharacter{1F004}{\private@AppleEmoji{1F004}}
\DeclareUnicodeCharacter{1F3B4}{\private@AppleEmoji{1F3B4}}
\DeclareUnicodeCharacter{1F507}{\private@AppleEmoji{1F507}}
\DeclareUnicodeCharacter{1F508}{\private@AppleEmoji{1F508}}
\DeclareUnicodeCharacter{1F509}{\private@AppleEmoji{1F509}}
\DeclareUnicodeCharacter{1F50A}{\private@AppleEmoji{1F50A}}
\DeclareUnicodeCharacter{1F4E2}{\private@AppleEmoji{1F4E2}}
\DeclareUnicodeCharacter{1F4E3}{\private@AppleEmoji{1F4E3}}
\DeclareUnicodeCharacter{1F4EF}{\private@AppleEmoji{1F4EF}}
\DeclareUnicodeCharacter{1F514}{\private@AppleEmoji{1F514}}
\DeclareUnicodeCharacter{1F515}{\private@AppleEmoji{1F515}}
\DeclareUnicodeCharacter{1F3BC}{\private@AppleEmoji{1F3BC}}
\DeclareUnicodeCharacter{1F3B5}{\private@AppleEmoji{1F3B5}}
\DeclareUnicodeCharacter{1F3B6}{\private@AppleEmoji{1F3B6}}
\DeclareUnicodeCharacter{1F399}{\private@AppleEmoji{1F399}}
\DeclareUnicodeCharacter{1F39A}{\private@AppleEmoji{1F39A}}
\DeclareUnicodeCharacter{1F39B}{\private@AppleEmoji{1F39B}}
\DeclareUnicodeCharacter{1F3A4}{\private@AppleEmoji{1F3A4}}
\DeclareUnicodeCharacter{1F3A7}{\private@AppleEmoji{1F3A7}}
\DeclareUnicodeCharacter{1F3B7}{\private@AppleEmoji{1F3B7}}
\DeclareUnicodeCharacter{1F3B8}{\private@AppleEmoji{1F3B8}}
\DeclareUnicodeCharacter{1F3B9}{\private@AppleEmoji{1F3B9}}
\DeclareUnicodeCharacter{1F3BA}{\private@AppleEmoji{1F3BA}}
\DeclareUnicodeCharacter{1F3BB}{\private@AppleEmoji{1F3BB}}
\DeclareUnicodeCharacter{1F4FB}{\private@AppleEmoji{1F4FB}}
\DeclareUnicodeCharacter{1F4F1}{\private@AppleEmoji{1F4F1}}
\DeclareUnicodeCharacter{1F4F2}{\private@AppleEmoji{1F4F2}}
\DeclareUnicodeCharacter{260E}{\private@AppleEmoji{260E}}
\DeclareUnicodeCharacter{1F4DE}{\private@AppleEmoji{1F4DE}}
\DeclareUnicodeCharacter{1F4DF}{\private@AppleEmoji{1F4DF}}
\DeclareUnicodeCharacter{1F4E0}{\private@AppleEmoji{1F4E0}}
\DeclareUnicodeCharacter{1F50B}{\private@AppleEmoji{1F50B}}
\DeclareUnicodeCharacter{1F50C}{\private@AppleEmoji{1F50C}}
\DeclareUnicodeCharacter{1F4BB}{\private@AppleEmoji{1F4BB}}
\DeclareUnicodeCharacter{1F5A5}{\private@AppleEmoji{1F5A5}}
\DeclareUnicodeCharacter{1F5A8}{\private@AppleEmoji{1F5A8}}
\DeclareUnicodeCharacter{2328}{\private@AppleEmoji{2328}}
\DeclareUnicodeCharacter{1F5B1}{\private@AppleEmoji{1F5B1}}
\DeclareUnicodeCharacter{1F5B2}{\private@AppleEmoji{1F5B2}}
\DeclareUnicodeCharacter{1F4BD}{\private@AppleEmoji{1F4BD}}
\DeclareUnicodeCharacter{1F4BE}{\private@AppleEmoji{1F4BE}}
\DeclareUnicodeCharacter{1F4BF}{\private@AppleEmoji{1F4BF}}
\DeclareUnicodeCharacter{1F4C0}{\private@AppleEmoji{1F4C0}}
\DeclareUnicodeCharacter{1F3A5}{\private@AppleEmoji{1F3A5}}
\DeclareUnicodeCharacter{1F3AC}{\private@AppleEmoji{1F3AC}}
\DeclareUnicodeCharacter{1F4FD}{\private@AppleEmoji{1F4FD}}
\DeclareUnicodeCharacter{1F4FA}{\private@AppleEmoji{1F4FA}}
\DeclareUnicodeCharacter{1F4F7}{\private@AppleEmoji{1F4F7}}
\DeclareUnicodeCharacter{1F4F8}{\private@AppleEmoji{1F4F8}}
\DeclareUnicodeCharacter{1F4F9}{\private@AppleEmoji{1F4F9}}
\DeclareUnicodeCharacter{1F4FC}{\private@AppleEmoji{1F4FC}}
\DeclareUnicodeCharacter{1F50D}{\private@AppleEmoji{1F50D}}
\DeclareUnicodeCharacter{1F50E}{\private@AppleEmoji{1F50E}}
\DeclareUnicodeCharacter{1F52C}{\private@AppleEmoji{1F52C}}
\DeclareUnicodeCharacter{1F52D}{\private@AppleEmoji{1F52D}}
\DeclareUnicodeCharacter{1F4E1}{\private@AppleEmoji{1F4E1}}
\DeclareUnicodeCharacter{1F56F}{\private@AppleEmoji{1F56F}}
\DeclareUnicodeCharacter{1F4A1}{\private@AppleEmoji{1F4A1}}
\DeclareUnicodeCharacter{1F526}{\private@AppleEmoji{1F526}}
\DeclareUnicodeCharacter{1F3EE}{\private@AppleEmoji{1F3EE}}
\DeclareUnicodeCharacter{1F4D4}{\private@AppleEmoji{1F4D4}}
\DeclareUnicodeCharacter{1F4D5}{\private@AppleEmoji{1F4D5}}
\DeclareUnicodeCharacter{1F4D6}{\private@AppleEmoji{1F4D6}}
\DeclareUnicodeCharacter{1F4D7}{\private@AppleEmoji{1F4D7}}
\DeclareUnicodeCharacter{1F4D8}{\private@AppleEmoji{1F4D8}}
\DeclareUnicodeCharacter{1F4D9}{\private@AppleEmoji{1F4D9}}
\DeclareUnicodeCharacter{1F4DA}{\private@AppleEmoji{1F4DA}}
\DeclareUnicodeCharacter{1F4D3}{\private@AppleEmoji{1F4D3}}
\DeclareUnicodeCharacter{1F4D2}{\private@AppleEmoji{1F4D2}}
\DeclareUnicodeCharacter{1F4C3}{\private@AppleEmoji{1F4C3}}
\DeclareUnicodeCharacter{1F4DC}{\private@AppleEmoji{1F4DC}}
\DeclareUnicodeCharacter{1F4C4}{\private@AppleEmoji{1F4C4}}
\DeclareUnicodeCharacter{1F4F0}{\private@AppleEmoji{1F4F0}}
\DeclareUnicodeCharacter{1F5DE}{\private@AppleEmoji{1F5DE}}
\DeclareUnicodeCharacter{1F4D1}{\private@AppleEmoji{1F4D1}}
\DeclareUnicodeCharacter{1F516}{\private@AppleEmoji{1F516}}
\DeclareUnicodeCharacter{1F4B0}{\private@AppleEmoji{1F4B0}}
\DeclareUnicodeCharacter{1F4B4}{\private@AppleEmoji{1F4B4}}
\DeclareUnicodeCharacter{1F4B5}{\private@AppleEmoji{1F4B5}}
\DeclareUnicodeCharacter{1F4B6}{\private@AppleEmoji{1F4B6}}
\DeclareUnicodeCharacter{1F4B7}{\private@AppleEmoji{1F4B7}}
\DeclareUnicodeCharacter{1F4B8}{\private@AppleEmoji{1F4B8}}
\DeclareUnicodeCharacter{1F4B3}{\private@AppleEmoji{1F4B3}}
\DeclareUnicodeCharacter{1F4B9}{\private@AppleEmoji{1F4B9}}
\DeclareUnicodeCharacter{2709}{\private@AppleEmoji{2709}}
\DeclareUnicodeCharacter{1F4E7}{\private@AppleEmoji{1F4E7}}
\DeclareUnicodeCharacter{1F4E8}{\private@AppleEmoji{1F4E8}}
\DeclareUnicodeCharacter{1F4E9}{\private@AppleEmoji{1F4E9}}
\DeclareUnicodeCharacter{1F4E4}{\private@AppleEmoji{1F4E4}}
\DeclareUnicodeCharacter{1F4E5}{\private@AppleEmoji{1F4E5}}
\DeclareUnicodeCharacter{1F4E6}{\private@AppleEmoji{1F4E6}}
\DeclareUnicodeCharacter{1F4EB}{\private@AppleEmoji{1F4EB}}
\DeclareUnicodeCharacter{1F4EA}{\private@AppleEmoji{1F4EA}}
\DeclareUnicodeCharacter{1F4EC}{\private@AppleEmoji{1F4EC}}
\DeclareUnicodeCharacter{1F4ED}{\private@AppleEmoji{1F4ED}}
\DeclareUnicodeCharacter{1F4EE}{\private@AppleEmoji{1F4EE}}
\DeclareUnicodeCharacter{1F5F3}{\private@AppleEmoji{1F5F3}}
\DeclareUnicodeCharacter{270F}{\private@AppleEmoji{270F}}
\DeclareUnicodeCharacter{2712}{\private@AppleEmoji{2712}}
\DeclareUnicodeCharacter{1F58B}{\private@AppleEmoji{1F58B}}
\DeclareUnicodeCharacter{1F58A}{\private@AppleEmoji{1F58A}}
\DeclareUnicodeCharacter{1F58C}{\private@AppleEmoji{1F58C}}
\DeclareUnicodeCharacter{1F58D}{\private@AppleEmoji{1F58D}}
\DeclareUnicodeCharacter{1F4DD}{\private@AppleEmoji{1F4DD}}
\DeclareUnicodeCharacter{1F4BC}{\private@AppleEmoji{1F4BC}}
\DeclareUnicodeCharacter{1F4C1}{\private@AppleEmoji{1F4C1}}
\DeclareUnicodeCharacter{1F4C2}{\private@AppleEmoji{1F4C2}}
\DeclareUnicodeCharacter{1F5C2}{\private@AppleEmoji{1F5C2}}
\DeclareUnicodeCharacter{1F4C5}{\private@AppleEmoji{1F4C5}}
\DeclareUnicodeCharacter{1F4C6}{\private@AppleEmoji{1F4C6}}
\DeclareUnicodeCharacter{1F5D2}{\private@AppleEmoji{1F5D2}}
\DeclareUnicodeCharacter{1F5D3}{\private@AppleEmoji{1F5D3}}
\DeclareUnicodeCharacter{1F4C7}{\private@AppleEmoji{1F4C7}}
\DeclareUnicodeCharacter{1F4C8}{\private@AppleEmoji{1F4C8}}
\DeclareUnicodeCharacter{1F4C9}{\private@AppleEmoji{1F4C9}}
\DeclareUnicodeCharacter{1F4CA}{\private@AppleEmoji{1F4CA}}
\DeclareUnicodeCharacter{1F4CB}{\private@AppleEmoji{1F4CB}}
\DeclareUnicodeCharacter{1F4CC}{\private@AppleEmoji{1F4CC}}
\DeclareUnicodeCharacter{1F4CD}{\private@AppleEmoji{1F4CD}}
\DeclareUnicodeCharacter{1F4CE}{\private@AppleEmoji{1F4CE}}
\DeclareUnicodeCharacter{1F587}{\private@AppleEmoji{1F587}}
\DeclareUnicodeCharacter{1F4CF}{\private@AppleEmoji{1F4CF}}
\DeclareUnicodeCharacter{1F4D0}{\private@AppleEmoji{1F4D0}}
\DeclareUnicodeCharacter{2702}{\private@AppleEmoji{2702}}
\DeclareUnicodeCharacter{1F5C3}{\private@AppleEmoji{1F5C3}}
\DeclareUnicodeCharacter{1F5C4}{\private@AppleEmoji{1F5C4}}
\DeclareUnicodeCharacter{1F5D1}{\private@AppleEmoji{1F5D1}}
\DeclareUnicodeCharacter{1F512}{\private@AppleEmoji{1F512}}
\DeclareUnicodeCharacter{1F513}{\private@AppleEmoji{1F513}}
\DeclareUnicodeCharacter{1F50F}{\private@AppleEmoji{1F50F}}
\DeclareUnicodeCharacter{1F510}{\private@AppleEmoji{1F510}}
\DeclareUnicodeCharacter{1F511}{\private@AppleEmoji{1F511}}
\DeclareUnicodeCharacter{1F5DD}{\private@AppleEmoji{1F5DD}}
\DeclareUnicodeCharacter{1F528}{\private@AppleEmoji{1F528}}
\DeclareUnicodeCharacter{26CF}{\private@AppleEmoji{26CF}}
\DeclareUnicodeCharacter{2692}{\private@AppleEmoji{2692}}
\DeclareUnicodeCharacter{1F6E0}{\private@AppleEmoji{1F6E0}}
\DeclareUnicodeCharacter{1F527}{\private@AppleEmoji{1F527}}
\DeclareUnicodeCharacter{1F529}{\private@AppleEmoji{1F529}}
\DeclareUnicodeCharacter{2699}{\private@AppleEmoji{2699}}
\DeclareUnicodeCharacter{1F5DC}{\private@AppleEmoji{1F5DC}}
\DeclareUnicodeCharacter{2697}{\private@AppleEmoji{2697}}
\DeclareUnicodeCharacter{2696}{\private@AppleEmoji{2696}}
\DeclareUnicodeCharacter{1F517}{\private@AppleEmoji{1F517}}
\DeclareUnicodeCharacter{26D3}{\private@AppleEmoji{26D3}}
\DeclareUnicodeCharacter{1F489}{\private@AppleEmoji{1F489}}
\DeclareUnicodeCharacter{1F48A}{\private@AppleEmoji{1F48A}}
\DeclareUnicodeCharacter{1F5E1}{\private@AppleEmoji{1F5E1}}
\DeclareUnicodeCharacter{1F52A}{\private@AppleEmoji{1F52A}}
\DeclareUnicodeCharacter{2694}{\private@AppleEmoji{2694}}
\DeclareUnicodeCharacter{1F52B}{\private@AppleEmoji{1F52B}}
\DeclareUnicodeCharacter{1F6E1}{\private@AppleEmoji{1F6E1}}
\DeclareUnicodeCharacter{1F3F9}{\private@AppleEmoji{1F3F9}}
\DeclareUnicodeCharacter{1F3C1}{\private@AppleEmoji{1F3C1}}
\DeclareUnicodeCharacter{1F3F3}{\private@AppleEmoji{1F3F3}}
\DeclareUnicodeCharacter{1F3F4}{\private@AppleEmoji{1F3F4}}
\DeclareUnicodeCharacter{1F6A9}{\private@AppleEmoji{1F6A9}}
\DeclareUnicodeCharacter{1F6AC}{\private@AppleEmoji{1F6AC}}
\DeclareUnicodeCharacter{26B0}{\private@AppleEmoji{26B0}}
\DeclareUnicodeCharacter{26B1}{\private@AppleEmoji{26B1}}
\DeclareUnicodeCharacter{1F5FF}{\private@AppleEmoji{1F5FF}}
\DeclareUnicodeCharacter{1F6E2}{\private@AppleEmoji{1F6E2}}
\DeclareUnicodeCharacter{1F52E}{\private@AppleEmoji{1F52E}}
\DeclareUnicodeCharacter{1F3E7}{\private@AppleEmoji{1F3E7}}
\DeclareUnicodeCharacter{1F6AE}{\private@AppleEmoji{1F6AE}}
\DeclareUnicodeCharacter{1F6B0}{\private@AppleEmoji{1F6B0}}
\DeclareUnicodeCharacter{267F}{\private@AppleEmoji{267F}}
\DeclareUnicodeCharacter{1F6B9}{\private@AppleEmoji{1F6B9}}
\DeclareUnicodeCharacter{1F6BA}{\private@AppleEmoji{1F6BA}}
\DeclareUnicodeCharacter{1F6BB}{\private@AppleEmoji{1F6BB}}
\DeclareUnicodeCharacter{1F6BC}{\private@AppleEmoji{1F6BC}}
\DeclareUnicodeCharacter{1F6BE}{\private@AppleEmoji{1F6BE}}
\DeclareUnicodeCharacter{1F6C2}{\private@AppleEmoji{1F6C2}}
\DeclareUnicodeCharacter{1F6C3}{\private@AppleEmoji{1F6C3}}
\DeclareUnicodeCharacter{1F6C4}{\private@AppleEmoji{1F6C4}}
\DeclareUnicodeCharacter{1F6C5}{\private@AppleEmoji{1F6C5}}
\DeclareUnicodeCharacter{26A0}{\private@AppleEmoji{26A0}}
\DeclareUnicodeCharacter{1F6B8}{\private@AppleEmoji{1F6B8}}
\DeclareUnicodeCharacter{26D4}{\private@AppleEmoji{26D4}}
\DeclareUnicodeCharacter{1F6AB}{\private@AppleEmoji{1F6AB}}
\DeclareUnicodeCharacter{1F6B3}{\private@AppleEmoji{1F6B3}}
\DeclareUnicodeCharacter{1F6AD}{\private@AppleEmoji{1F6AD}}
\DeclareUnicodeCharacter{1F6AF}{\private@AppleEmoji{1F6AF}}
\DeclareUnicodeCharacter{1F6B1}{\private@AppleEmoji{1F6B1}}
\DeclareUnicodeCharacter{1F6B7}{\private@AppleEmoji{1F6B7}}
\DeclareUnicodeCharacter{2622}{\private@AppleEmoji{2622}}
\DeclareUnicodeCharacter{2623}{\private@AppleEmoji{2623}}
\DeclareUnicodeCharacter{2B06}{\private@AppleEmoji{2B06}}
\DeclareUnicodeCharacter{2197}{\private@AppleEmoji{2197}}
\DeclareUnicodeCharacter{27A1}{\private@AppleEmoji{27A1}}
\DeclareUnicodeCharacter{2198}{\private@AppleEmoji{2198}}
\DeclareUnicodeCharacter{2B07}{\private@AppleEmoji{2B07}}
\DeclareUnicodeCharacter{2199}{\private@AppleEmoji{2199}}
\DeclareUnicodeCharacter{2B05}{\private@AppleEmoji{2B05}}
\DeclareUnicodeCharacter{2196}{\private@AppleEmoji{2196}}
\DeclareUnicodeCharacter{2195}{\private@AppleEmoji{2195}}
\DeclareUnicodeCharacter{2194}{\private@AppleEmoji{2194}}
\DeclareUnicodeCharacter{21A9}{\private@AppleEmoji{21A9}}
\DeclareUnicodeCharacter{21AA}{\private@AppleEmoji{21AA}}
\DeclareUnicodeCharacter{2934}{\private@AppleEmoji{2934}}
\DeclareUnicodeCharacter{2935}{\private@AppleEmoji{2935}}
\DeclareUnicodeCharacter{1F503}{\private@AppleEmoji{1F503}}
\DeclareUnicodeCharacter{1F504}{\private@AppleEmoji{1F504}}
\DeclareUnicodeCharacter{1F519}{\private@AppleEmoji{1F519}}
\DeclareUnicodeCharacter{1F51A}{\private@AppleEmoji{1F51A}}
\DeclareUnicodeCharacter{1F51B}{\private@AppleEmoji{1F51B}}
\DeclareUnicodeCharacter{1F51C}{\private@AppleEmoji{1F51C}}
\DeclareUnicodeCharacter{1F51D}{\private@AppleEmoji{1F51D}}
\DeclareUnicodeCharacter{1F6D0}{\private@AppleEmoji{1F6D0}}
\DeclareUnicodeCharacter{269B}{\private@AppleEmoji{269B}}
\DeclareUnicodeCharacter{1F549}{\private@AppleEmoji{1F549}}
\DeclareUnicodeCharacter{2721}{\private@AppleEmoji{2721}}
\DeclareUnicodeCharacter{2638}{\private@AppleEmoji{2638}}
\DeclareUnicodeCharacter{262F}{\private@AppleEmoji{262F}}
\DeclareUnicodeCharacter{271D}{\private@AppleEmoji{271D}}
\DeclareUnicodeCharacter{2626}{\private@AppleEmoji{2626}}
\DeclareUnicodeCharacter{262A}{\private@AppleEmoji{262A}}
\DeclareUnicodeCharacter{262E}{\private@AppleEmoji{262E}}
\DeclareUnicodeCharacter{1F54E}{\private@AppleEmoji{1F54E}}
\DeclareUnicodeCharacter{1F52F}{\private@AppleEmoji{1F52F}}
\DeclareUnicodeCharacter{267B}{\private@AppleEmoji{267B}}
\DeclareUnicodeCharacter{1F4DB}{\private@AppleEmoji{1F4DB}}
\DeclareUnicodeCharacter{269C}{\private@AppleEmoji{269C}}
\DeclareUnicodeCharacter{1F530}{\private@AppleEmoji{1F530}}
\DeclareUnicodeCharacter{1F531}{\private@AppleEmoji{1F531}}
\DeclareUnicodeCharacter{2B55}{\private@AppleEmoji{2B55}}
\DeclareUnicodeCharacter{2705}{\private@AppleEmoji{2705}}
\DeclareUnicodeCharacter{2611}{\private@AppleEmoji{2611}}
\DeclareUnicodeCharacter{2714}{\private@AppleEmoji{2714}}
\DeclareUnicodeCharacter{2716}{\private@AppleEmoji{2716}}
\DeclareUnicodeCharacter{274C}{\private@AppleEmoji{274C}}
\DeclareUnicodeCharacter{274E}{\private@AppleEmoji{274E}}
\DeclareUnicodeCharacter{2795}{\private@AppleEmoji{2795}}
\DeclareUnicodeCharacter{2796}{\private@AppleEmoji{2796}}
\DeclareUnicodeCharacter{2797}{\private@AppleEmoji{2797}}
\DeclareUnicodeCharacter{27B0}{\private@AppleEmoji{27B0}}
\DeclareUnicodeCharacter{27BF}{\private@AppleEmoji{27BF}}
\DeclareUnicodeCharacter{303D}{\private@AppleEmoji{303D}}
\DeclareUnicodeCharacter{2733}{\private@AppleEmoji{2733}}
\DeclareUnicodeCharacter{2734}{\private@AppleEmoji{2734}}
\DeclareUnicodeCharacter{2747}{\private@AppleEmoji{2747}}
\DeclareUnicodeCharacter{1F4B1}{\private@AppleEmoji{1F4B1}}
\DeclareUnicodeCharacter{1F4B2}{\private@AppleEmoji{1F4B2}}
\DeclareUnicodeCharacter{203C}{\private@AppleEmoji{203C}}
\DeclareUnicodeCharacter{2049}{\private@AppleEmoji{2049}}
\DeclareUnicodeCharacter{2753}{\private@AppleEmoji{2753}}
\DeclareUnicodeCharacter{2754}{\private@AppleEmoji{2754}}
\DeclareUnicodeCharacter{2755}{\private@AppleEmoji{2755}}
\DeclareUnicodeCharacter{2757}{\private@AppleEmoji{2757}}
\DeclareUnicodeCharacter{3030}{\private@AppleEmoji{3030}}
\DeclareUnicodeCharacter{00A9}{\private@AppleEmoji{00A9}}
\DeclareUnicodeCharacter{00AE}{\private@AppleEmoji{00AE}}
\DeclareUnicodeCharacter{2122}{\private@AppleEmoji{2122}}
\DeclareUnicodeCharacter{2648}{\private@AppleEmoji{2648}}
\DeclareUnicodeCharacter{2649}{\private@AppleEmoji{2649}}
\DeclareUnicodeCharacter{264A}{\private@AppleEmoji{264A}}
\DeclareUnicodeCharacter{264B}{\private@AppleEmoji{264B}}
\DeclareUnicodeCharacter{264C}{\private@AppleEmoji{264C}}
\DeclareUnicodeCharacter{264D}{\private@AppleEmoji{264D}}
\DeclareUnicodeCharacter{264E}{\private@AppleEmoji{264E}}
\DeclareUnicodeCharacter{264F}{\private@AppleEmoji{264F}}
\DeclareUnicodeCharacter{2650}{\private@AppleEmoji{2650}}
\DeclareUnicodeCharacter{2651}{\private@AppleEmoji{2651}}
\DeclareUnicodeCharacter{2652}{\private@AppleEmoji{2652}}
\DeclareUnicodeCharacter{2653}{\private@AppleEmoji{2653}}
\DeclareUnicodeCharacter{26CE}{\private@AppleEmoji{26CE}}
\DeclareUnicodeCharacter{1F500}{\private@AppleEmoji{1F500}}
\DeclareUnicodeCharacter{1F501}{\private@AppleEmoji{1F501}}
\DeclareUnicodeCharacter{1F502}{\private@AppleEmoji{1F502}}
\DeclareUnicodeCharacter{25B6}{\private@AppleEmoji{25B6}}
\DeclareUnicodeCharacter{23E9}{\private@AppleEmoji{23E9}}
\DeclareUnicodeCharacter{23ED}{\private@AppleEmoji{23ED}}
\DeclareUnicodeCharacter{23EF}{\private@AppleEmoji{23EF}}
\DeclareUnicodeCharacter{25C0}{\private@AppleEmoji{25C0}}
\DeclareUnicodeCharacter{23EA}{\private@AppleEmoji{23EA}}
\DeclareUnicodeCharacter{23EE}{\private@AppleEmoji{23EE}}
\DeclareUnicodeCharacter{1F53C}{\private@AppleEmoji{1F53C}}
\DeclareUnicodeCharacter{23EB}{\private@AppleEmoji{23EB}}
\DeclareUnicodeCharacter{1F53D}{\private@AppleEmoji{1F53D}}
\DeclareUnicodeCharacter{23EC}{\private@AppleEmoji{23EC}}
\DeclareUnicodeCharacter{23F8}{\private@AppleEmoji{23F8}}
\DeclareUnicodeCharacter{23F9}{\private@AppleEmoji{23F9}}
\DeclareUnicodeCharacter{23FA}{\private@AppleEmoji{23FA}}
\DeclareUnicodeCharacter{1F3A6}{\private@AppleEmoji{1F3A6}}
\DeclareUnicodeCharacter{1F505}{\private@AppleEmoji{1F505}}
\DeclareUnicodeCharacter{1F506}{\private@AppleEmoji{1F506}}
\DeclareUnicodeCharacter{1F4F6}{\private@AppleEmoji{1F4F6}}
\DeclareUnicodeCharacter{1F4F5}{\private@AppleEmoji{1F4F5}}
\DeclareUnicodeCharacter{1F4F3}{\private@AppleEmoji{1F4F3}}
\DeclareUnicodeCharacter{1F4F4}{\private@AppleEmoji{1F4F4}}
\DeclareUnicodeCharacter{1F51F}{\private@AppleEmoji{1F51F}}
\DeclareUnicodeCharacter{1F4AF}{\private@AppleEmoji{1F4AF}}
\DeclareUnicodeCharacter{1F51E}{\private@AppleEmoji{1F51E}}
\DeclareUnicodeCharacter{1F520}{\private@AppleEmoji{1F520}}
\DeclareUnicodeCharacter{1F521}{\private@AppleEmoji{1F521}}
\DeclareUnicodeCharacter{1F522}{\private@AppleEmoji{1F522}}
\DeclareUnicodeCharacter{1F523}{\private@AppleEmoji{1F523}}
\DeclareUnicodeCharacter{1F524}{\private@AppleEmoji{1F524}}
\DeclareUnicodeCharacter{1F170}{\private@AppleEmoji{1F170}}
\DeclareUnicodeCharacter{1F18E}{\private@AppleEmoji{1F18E}}
\DeclareUnicodeCharacter{1F171}{\private@AppleEmoji{1F171}}
\DeclareUnicodeCharacter{1F191}{\private@AppleEmoji{1F191}}
\DeclareUnicodeCharacter{1F192}{\private@AppleEmoji{1F192}}
\DeclareUnicodeCharacter{1F193}{\private@AppleEmoji{1F193}}
\DeclareUnicodeCharacter{2139}{\private@AppleEmoji{2139}}
\DeclareUnicodeCharacter{1F194}{\private@AppleEmoji{1F194}}
\DeclareUnicodeCharacter{24C2}{\private@AppleEmoji{24C2}}
\DeclareUnicodeCharacter{1F195}{\private@AppleEmoji{1F195}}
\DeclareUnicodeCharacter{1F196}{\private@AppleEmoji{1F196}}
\DeclareUnicodeCharacter{1F17E}{\private@AppleEmoji{1F17E}}
\DeclareUnicodeCharacter{1F197}{\private@AppleEmoji{1F197}}
\DeclareUnicodeCharacter{1F17F}{\private@AppleEmoji{1F17F}}
\DeclareUnicodeCharacter{1F198}{\private@AppleEmoji{1F198}}
\DeclareUnicodeCharacter{1F199}{\private@AppleEmoji{1F199}}
\DeclareUnicodeCharacter{1F19A}{\private@AppleEmoji{1F19A}}
\DeclareUnicodeCharacter{1F201}{\private@AppleEmoji{1F201}}
\DeclareUnicodeCharacter{1F202}{\private@AppleEmoji{1F202}}
\DeclareUnicodeCharacter{1F237}{\private@AppleEmoji{1F237}}
\DeclareUnicodeCharacter{1F236}{\private@AppleEmoji{1F236}}
\DeclareUnicodeCharacter{1F22F}{\private@AppleEmoji{1F22F}}
\DeclareUnicodeCharacter{1F250}{\private@AppleEmoji{1F250}}
\DeclareUnicodeCharacter{1F239}{\private@AppleEmoji{1F239}}
\DeclareUnicodeCharacter{1F21A}{\private@AppleEmoji{1F21A}}
\DeclareUnicodeCharacter{1F232}{\private@AppleEmoji{1F232}}
\DeclareUnicodeCharacter{1F251}{\private@AppleEmoji{1F251}}
\DeclareUnicodeCharacter{1F238}{\private@AppleEmoji{1F238}}
\DeclareUnicodeCharacter{1F234}{\private@AppleEmoji{1F234}}
\DeclareUnicodeCharacter{1F233}{\private@AppleEmoji{1F233}}
\DeclareUnicodeCharacter{3297}{\private@AppleEmoji{3297}}
\DeclareUnicodeCharacter{3299}{\private@AppleEmoji{3299}}
\DeclareUnicodeCharacter{1F23A}{\private@AppleEmoji{1F23A}}
\DeclareUnicodeCharacter{1F235}{\private@AppleEmoji{1F235}}
\DeclareUnicodeCharacter{25AA}{\private@AppleEmoji{25AA}}
\DeclareUnicodeCharacter{25AB}{\private@AppleEmoji{25AB}}
\DeclareUnicodeCharacter{25FB}{\private@AppleEmoji{25FB}}
\DeclareUnicodeCharacter{25FC}{\private@AppleEmoji{25FC}}
\DeclareUnicodeCharacter{25FD}{\private@AppleEmoji{25FD}}
\DeclareUnicodeCharacter{25FE}{\private@AppleEmoji{25FE}}
\DeclareUnicodeCharacter{2B1B}{\private@AppleEmoji{2B1B}}
\DeclareUnicodeCharacter{2B1C}{\private@AppleEmoji{2B1C}}
\DeclareUnicodeCharacter{1F536}{\private@AppleEmoji{1F536}}
\DeclareUnicodeCharacter{1F537}{\private@AppleEmoji{1F537}}
\DeclareUnicodeCharacter{1F538}{\private@AppleEmoji{1F538}}
\DeclareUnicodeCharacter{1F539}{\private@AppleEmoji{1F539}}
\DeclareUnicodeCharacter{1F53A}{\private@AppleEmoji{1F53A}}
\DeclareUnicodeCharacter{1F53B}{\private@AppleEmoji{1F53B}}
\DeclareUnicodeCharacter{1F4A0}{\private@AppleEmoji{1F4A0}}
\DeclareUnicodeCharacter{1F518}{\private@AppleEmoji{1F518}}
\DeclareUnicodeCharacter{1F532}{\private@AppleEmoji{1F532}}
\DeclareUnicodeCharacter{1F533}{\private@AppleEmoji{1F533}}
\DeclareUnicodeCharacter{26AA}{\private@AppleEmoji{26AA}}
\DeclareUnicodeCharacter{26AB}{\private@AppleEmoji{26AB}}
\DeclareUnicodeCharacter{1F534}{\private@AppleEmoji{1F534}}
\DeclareUnicodeCharacter{1F535}{\private@AppleEmoji{1F535}}
\makeatother

\newcolumntype{L}[1]{>{\raggedright\let\newline\\\arraybackslash\hspace{0pt}}m{#1}}
\newcolumntype{C}[1]{>{\centering\let\newline\\\arraybackslash\hspace{0pt}}m{#1}}
\newcolumntype{R}[1]{>{\raggedleft\let\newline\\\arraybackslash\hspace{0pt}}m{#1}}
\usepackage{colortbl}
\definecolor{LGray}{gray}{0.9}
\definecolor{Gray}{gray}{0.8}
\definecolor{DGray}{gray}{0.7}
\makeatother

\title{Studying Cultural Differences in Emoji Usage across the East and the West}
\author{Sharath Chandra Guntuku$^{1}$, Mingyang Li$^{1}$, Louis Tay$^{2}$, Lyle H. Ungar$^{1}$\\
       $^{1}$University of Pennsylvania, $^{2}$Purdue University\\
       \{sharathg@sas, myli@seas, ungar@cis\}.upenn.edu, stay@purdue.edu
       }
\begin{document}

\maketitle
\begin{abstract}
Global acceptance of Emojis suggests a cross-cultural, normative use
of Emojis. Meanwhile, nuances in Emoji use across cultures may
also exist due to linguistic differences in expressing emotions and
diversity in conceptualizing topics. Indeed, literature in cross-cultural
psychology has found both normative and culture-specific ways in which
emotions are expressed. In this paper, using social media, we compare
the Emoji usage based on frequency, context, and topic associations
across countries in the East (China and Japan) and the West (United
States, United Kingdom, and Canada). 
Across the East and the West,
our study examines a) similarities and differences on the usage of
different categories of Emojis such as People, Food \& Drink, Travel
\& Places etc., b) potential mapping of Emoji
use differences with previously identified cultural differences in
users' expression about diverse concepts such as death, money emotions
and family, and c) relative correspondence of validated psycho-linguistic
categories with Ekman's emotions.
The analysis of Emoji use in the East and the West reveals
recognizable normative and culture specific patterns. This research
reveals the ways in which Emojis can be used for cross-cultural communication. 
\end{abstract}

\section{Introduction}

Emoji, a Japan-born ideographic system, offers a rich set of non-verbal
cues to assist textual communication. The Unicode Standard 11.0 specified
over $2,500$ Emojis\footnote{\url{http://unicode.org/Emoji/}}, ranging
from facial expressions (`Smileys' such as 😄) to everyday objects
(such as 🚘). Starting as a visual aid for textual communication, Emojis'
non-verbal nature has led to suggestions that they are universal across
cultures~\cite{danesi2016semiotics}. In this paper, we examine cross-cultural
usages of Emojis based on (1) linguistic differences across languages
(and cultures) in expressing emotions~\cite{russell2013everyday},
and (2) diversity in perceiving different constructs among cultures~\cite{boers2003applied}.
Specifically, we compare Emoji use in terms of frequency,
context, and topic associations across two eastern countries -- China
and Japan -- and three western countries -- United States of America
(US), United Kingdom (UK) and Canada. Hereafter, we refer to the collection
of US, UK, and Canadian cultures as `the Western culture' (or simply
`the West'), the collection of Japanese and Mandarin-speaking Chinese
as `the East(-ern culture)' with an acknowledgment that there are
several other countries which can be added to each group~\cite{hofstede1983national}.
We study the differences using distributional semantics learned over
large datasets from Sina Weibo and Twitter, two closely related microblog/social media platforms. 

Past psychological research assessing emotional experience between
the East and the West found both universal and culture-specific types
of emotional experience~\cite{eid2009norms}. If Emojis are a form
of emotional experience and expression as prior studies have shown~\cite{kaye2017emojis}, it is expected that we can find interpretable and substantial similarity in Emoji usage and also distinct cultural patterns. In other words, which Emojis are used, the contexts where they are used, and what they semantically refer to will bear resemblance across languages, even when no common character is shared between their writing systems. At the same time, there will also be unique cultural elements in how
certain types of Emojis are used and interpreted.

\subsection{Research Questions}

Due to the richness and diversity of the Emojis, it is difficult to
hypothesize \textit{a priori} how specific Emojis may differ. Therefore,
we undertake an abductive approach to construct explanatory theories
as patterns emerge from our analysis~\cite{haig2005abductive}. Normatively,
we fundamentally expect similar patterns of Emoji usage to appear
across both cultures. We also seek to explore when there may be specific cultural divergence. Therefore, we attempt to answer the following
research questions to explore and quantify the normativeness and distinctiveness
of Emoji usage across the two cultures:
\begin{enumerate}
\item How does frequency of different Emojis (as individuals and in categories)
vary across the East and the West? 
\item How distinct is Emoji usage across cultures in terms of validated
psycho-linguistic categories they are often associated with? 
\item How do the semantics of Emoji usage vary when compared against known
universal emotion expressions emotions (Ekman basic emotion categories~\cite{Ekman1992argument})
across both cultures? 
\end{enumerate}

\section{Background and Related Work}

\subsection{Weibo and Twitter: Analogs?}
Weibo and Twitter have been studied to understand the differences
in content and user behaviors in multiple contexts~\cite{ma2013electronic,lin2016exploring}.
Notwithstanding the challenge of working with a non-random, non-representative
sample of social media users, several psychological traits and outcomes
can be inferred from posts, including users' demographics~\cite{sap2014developing,zhang2016predicting},
personality~\cite{li2014predicting,quercia2011our}, location~\cite{salehi2017huntsville,zhong2015you},
as well as status of stress~\cite{guntuku2018understanding,lin2016does}, and mental health~\cite{guntuku2019imagedep,tian2018characterizing}
on both platforms. Demonstrated in these prior studies, the empirical
value suggests that the two platforms are representative albeit to
different countries. Several of the above mentioned studies have used
Linguistic Inquiry Word Count (LIWC) \cite{pennebaker2015development},
which has psychometrically validated categories, such as
emotional valence, religion, and money etc. in multiple languages. 

\subsection{Role of Culture in Emotion Expression and Perception}
Prior psychological research on emotion~\cite{uchida2009happiness,bagozzi1999role}
suggests that evolutionary and biological processes generate universal
expressions and perceptions of emotions. For example, facial expressions
is one of such universal channels that convey emotions across populations~\cite{Ekman1993facial}.
On the other hand, culture can play a significant role in shaping
emotional life. Specifically, different cultures may value different
types of emotions (e.g., Americans value excitement while Asians prefer
calm)~\cite{tsai2006cultural}, and there are different emotional
display rules across cultures~\cite{matsumoto1990cultural}. Furthermore,
besides representing emotions, the Emoji system also explicitly contains
cultural symbols and thus potentially represents the distinctive values and beliefs
of cultures~\cite{aaker2001consumption}. Prior works also suggest that culture plays a key role in predicting perceptions of affect~\cite{guntuku2015cp,zhu2018measuring,guntuku2015modelling}. In general, psychological
research reveals both cultural similarities and differences in emotions~\cite{elfenbein2003universals}.

\subsection{Emojis: A Proxy for Emotions?}
From a methodological perspective, most large scale cross-cultural
psychology research projects have used a survey-based approach to
assess similarities and differences in emotions~\cite{tay2011multilevel}.
Analyzing use of Emojis between the Eastern and the Western contexts
provides an opportunity to assess a plethora of behaviors related
to emotion expression and, arguably, emotional symbols that are culturally
embedded. This enhances our understanding of similarities and differences
in emotional life across cultures at a large scale but fine-grain
level.

\subsection{Prior Studies on Emojis}

Prior research on Emojis can be divided primarily into three themes:
(1) studying Emojis as a source of sentiment annotation, (2) analyzing
differences in Emoji perception based on rendering, and (3) understanding
similarities and differences in Emoji usage across different populations.
We focus on studying the similarities and differences in Emoji usage
across cultures.

The studies by~\cite{barbieri2016cosmopolitan,barbieri2016revealing}
are the closest to our work where authors explore the meaning and
usage of Emojis in social media across four European languages, namely
American English, British English, Peninsular Spanish and Italian,
and across two cities in Spain respectively. They observe that the most frequent
Emojis share similar semantic usages across these Western languages
providing support for the normative claims of Emojis. However, to
fully examine the issue of normativeness, we need to go beyond examining
only Western contexts by also examining East-West similarity.

Since Emojis have been found to be very promising in downstream applications
such as sentiment analysis and several techniques, utilizing Unicode
descriptions~\cite{eisner2016Emoji2vec,wijeratne2017emojinet},
multi-Emoji expressions~\cite{lopez2017did}, and including diverse Emoji
sets~\cite{felbo2017using} etc., have been proposed to improve Emoji
understanding, culture-specific norms and platform-rending effects
\cite{li2019emoji,miller2018see,miller2016blissfully} can be used
for improving personalization.

\section{Methods}
This work received approval from University of Pennsylvania's Institutional Review Board
(IRB). Code repository is available online\footnote{\url{https://github.com/tslmy/ICWSM2019}}.

\subsection{Data Collection}
To obtain data for the US, UK, Canada and Japan, we used Twitter data from a $10\%$ archive from the TrendMiner
project~\cite{preotiuc2012trendminer}, which used the Twitter streaming
API.
Since Twitter is not widely used in China, we obtained Weibo data.
Since Weibo lacks a streaming interface (as Twitter) for downloading
random samples over time, we queried for all posts from a given user.
The list of users were crawled using a breadth-first search strategy
beginning with random users.

\begin{table*}
\center
\small
\begin{tabular}{|c|c|c|c|c|c|}
\hline 
\textbf{Culture}  & \textbf{Country}  & \textbf{\# Posts crawled}  & \textbf{\# Posts after lang. filter}  & \textbf{\# Posts after geo-location} & \textbf{\# Users} \tabularnewline
\hline 
\multirow{3}{*}{West } & USA  & 29.32M & 18.99M & 18.57M  & 4.39M \tabularnewline
\cline{2-2} \cline{3-6} 
 & UK  & 6.74M & 5.12M & 4.83M  & 1.31M \tabularnewline
\cline{2-2} \cline{3-6} 
 & Canada  & 1.6M & 1.16M & 1.12M  & 0.32M \tabularnewline
\hline 
\multirow{2}{*}{East} & Japan  & 481.39M  & 82.56M  & 17.51M  & 2.06M \tabularnewline
\cline{2-6} 
 & China  & 486.18M  & \multicolumn{2}{c|}{205.22M} & 1.00M \tabularnewline
\hline 
\end{tabular}\caption{Number of posts in each corpora after each pre-processing stage and
the final number of users in our analysis.}
\label{tab:cnts}
\end{table*}

\subsection{Pre-processing}

The count of posts for each country after each stage of pre-processing
and final user counts is shown in Table \ref{tab:cnts}.

\subsubsection{Geo-location:}

On Twitter, the coordinates or tweet country location (whichever
was available) was used to geo-locate posts. On Weibo, user’s self-identified
profile location was used to identify the geo-location of messages.
We used messages posted in the year 2014 in both corpora.

\subsubsection{Language Filtering:}

To remove the confounds of bilingualism~\cite{fishman1980bilingualism},
we filter posts by the languages they are composed in. Language used
in each Twitter post (or `tweets' hereafter) is detected via \textit{langid}~\cite{lui2012langid}.
Tweets written in languages other than English in US, UK and Canada,
and Japanese in Japan are removed. Weibo posts are filtered for Chinese
language using pre-trained \textit{fastText} language detection models
\cite{xu2011fast}, due to its ability to distinguish between Mandarin
and Cantonese (we used only Mandarin posts in our analysis). Further,
traditional Chinese characters are converted to Simplified Chinese
using \textit{hanziconv} Python package \footnote{\url{https://pypi.org/project/hanziconv/}}
to conform with LIWC dictionary used in later sections. We also remove
any direct re-tweets (indicated by `\texttt{RT @USERNAME:}' on Twitter
and `\texttt{@USERNAME//}' on Weibo).

\subsubsection{Tokenizing:}

Twitter text was tokenized using \textit{Social Tokenizer} bundled
with \textit{ekphrasis}\footnote{\url{https://github.com/cbaziotis/ekphrasis}},
a text processing pipeline designed for social networks. Weibo posts
were segmented using \textit{Jieba}\footnote{\url{https://github.com/fxsjy/jieba}}
considering its ability to discover new words and Internet slang,
which is particularly important for a highly colloquial corpus like
Sina Weibo. Using \textit{ekphrasis}, URLs, email addresses, percentages,
currency amounts, phone numbers, user names, emoticons and time-and-dates
were normalized with meta-tokens such as `\texttt{<url>}', `\texttt{<email>}',
`\texttt{<user>}' etc. Skin tone variation in Emojis was introduced
in 2015, and consequently no skin-toned Emoji was captured in our
corpora gathered from 2014.

\subsection{Training Embedding Models}

To study the lexical semantics across both cultures, we trained a
\textit{Word2Vec} Continuous Bag-of-Words (CBoW) model on each corpus/country~\cite{mikolov2013distributed}. These models were trained for 10
epochs with learning rates initialized at $.025$ and allowed to drop
till $10^{-4}$. The dimension of learned token vectors was chosen
to be $100$ based on previous work~\cite{barbieri2016does}. To
counter effects due to the randomized initialization in the \textit{Word2Vec}
algorithm, each model was trained 5 times independently. In all our
analysis, we used the vector embeddings across the 5 instances for
every analysis and then averaged the resulting projections.

\subsection{Measuring Topical Differences}
In order to investigate topical differences across cultures, we use
LIWC dictionaries in Chinese~\cite{huang2012development} and in
English ~\cite{pennebaker2001linguistic} to be consistent across
languages. Since LIWC is not available in Japanese, we used methods
from prior work~\cite{shibata2016detecting} to translate the word lists from Chinese and English into Japanese. The LIWC dictionary is a language-specific, many-to-many mapping of
tokens (including words and word stems) and psychologically validated
categories. Each category (a curated list of words)
is found to be correlated with and also predictive of several psychological traits and outcomes~\cite{pennebaker2001linguistic}.

We use the terms `tokens', `Emojis',
`words', etc. interchangeably with their corresponding vectorial representations. Next, we define our auxiliary term `category vectors', compute Emoji-category similarities, and analyze correlations.

\subsubsection{Preparing category vectors:}

In each corpus separately, for each LIWC category $i\in\left\{ \text{Posemo},\text{Family},...\right\} $,
all tokens (in vectorial representations $\vec{t}_{l,i_{j}}$) in
this category are averaged into one vector, which we term as ``category vector'' $\vec{c}_{l,i}$: 
\small{\[
\vec{c}_{l,i}=\frac{1}{n_{l,i}}\sum_{j=1}^{n_{l,i}}\vec{t}_{l,i_{j}}
\]}
for corpus $l\in\left\{ \text{US},\text{UK},\text{Canada},\text{Japan},\text{China}\right\} $
where $n_{l,i}$ is the amount of tokens in the LIWC category $i$
in the corpus $l$, and $\vec{t}_{l,i_{j}}$ is the $j$-th token
in the LIWC category $i$ in the corpus $l$.

Acknowledging that LIWC captures only verbal tokens, and that Emojis,
as non-verbal tokens, may have substantial differences to verbal tokens
captured during \textit{Word2Vec} training stage, we orthonormalize
axial vectors using Gram-Schmidt algorithm~\cite{bjorck1994numerics},
to ensures they capture more distinctive features
between LIWC categories.

\subsubsection{Computing cosine similarities:}

Separately in each corpus $l$, for each pair of Emoji $j\in\left\{ \right.$😄$,$
🚘$\left.,\ldots\right\} $ (in vectorial representation $\vec{t}_{l,j}$)
and category vector $\vec{c}_{l,i}$, a cosine similarity is computed:
\small{
\[
s_{l,i,j}=\text{sim}\left(\vec{c}_{l,i},\vec{t}_{l,j}\right).
\]}
For clarity, we define $\vec{s}_{l,i}=\left\{ s_{l,i,j}\text{ for }\forall j\right\} $.
Per-country cosine similarities are then averaged across each culture
to reveal the western and the eastern cosine similarities: 
\small{\[
\vec{s}_{W,i}=\frac{1}{3}\sum_{l\in\left\{ \text{US},\text{UK},\text{Canada}\right\} }\vec{s}_{l,i},
\]}
 and 
\small{\[
\vec{s}_{E,i}=\frac{1}{2}\sum_{l\in\left\{ \text{China},\text{Japan}\right\} }\vec{s}_{l,i}
\]}
 for the category $i\in\left\{ \text{Posemo},\text{Family},...\right\} $.

\subsubsection{Spearman Correlation Coefficients:}

For each of the $31$ LIWC categories shared across all corpora, \small{$i\in\left\{ \text{Posemo},\text{Family},...\right\} $},
we correlate the Western Emoji Usage vector \small{$\vec{s}_{\text{W},i}=\left(s_{\text{W},1},s_{\text{W},2},...,s_{\text{W},J}\right)$}
and the Eastern Emoji Usage vector \small{$\vec{s}_{\text{E},i}=\left(s_{\text{E},1},s_{\text{E},2},...,s_{\text{E},J}\right)$}, denoting the Spearman correlation coefficient with $\rho_{i}$. Here,
$J$ is the total number of Emojis present in all corpus and appeared
for at least $1,000$ times in total.

\section{Results and Discussion}

\subsection{Frequency of Emoji Usage}

\begin{figure}[t!]
	\center
	\includegraphics[width=1\columnwidth]{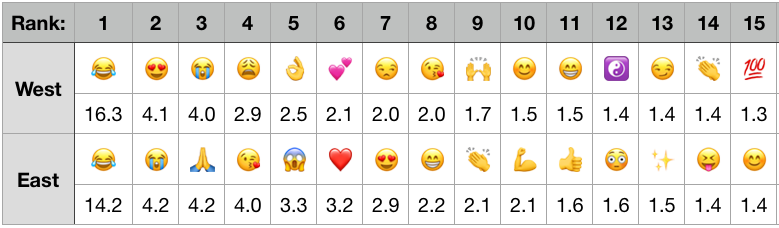}
	\caption{Top 15 frequent Emojis in the East and in the West, in percentage of total Emojis captured in the corresponding corpora. East-West rank order correlation is $.745$.}
	\label{fig:top20} 
\end{figure}

Among the $1,281$ Emojis defined in Emoji 1.0\footnote{Published in August 2015, Emoji 1.0 is closest to 2014, the year from
which our corpora were gathered.} by Unicode\footnote{\url{http://unicode.org/Public/emoji/1.0/emoji-data.txt}},
$602$ Emojis appeared in all corpora. Only $528$ of them appeared
more than $1,000$ times. Figure \ref{fig:top20} shows $15$ most
frequently seen Emojis in each culture. Across the two cultures, Spearman
correlation coefficient (SCC) is $0.745$ (two-tailed
t-test p-value $<0.005$). These statistics indicate a strong correspondence
in the types of Emojis favored  across these two cultures. This reveals
normativeness in the types of Emojis used between East and West.

\begin{figure}[t!]
	\centering{}\includegraphics[width=1\columnwidth]{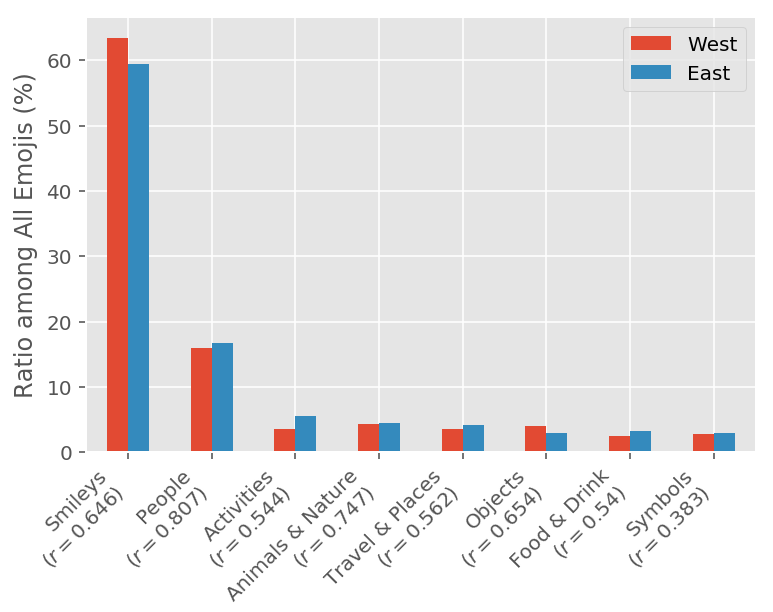}
	\caption{Normalized frequency of Emojis grouped by Unicode categories. SCCs across East and West, $r$, are denoted below each.}
	\label{fig:fig2} 
\end{figure}

Further, we sum up the usage frequencies by Unicode Category of Emojis.
Figure \ref{fig:fig2} shows the frequency of these categories, denoted
with SCCs for Emoji frequencies within each category, representing the similarity between Westerners and Easterners in using the Emojis in each category.
While the SCC values range from moderate ($.383$) to high ($.807$),
suggesting a high correspondence of Emoji usage patterns, drilling
down by categories of Emojis is elucidating. The lowest correlations
in Emoji usage occur in the `Symbols', `Food \& Drink', and `Activities' categories.
This is not surprising, because cultures often have their own meaning
symbols that are representative of specific values~\cite{aaker2001consumption}.
Moreover, culture is often instantiated in cuisines representing dietary
preferences, identities, and ecology~\cite{van1984ethnic}. Further, culture also influences the time spent across the world on work, play, and development activities~\cite{larson1999children}.

\subsection{Semantic Similarity of Emojis}

\begin{figure}[t!]
\centering{}\includegraphics[width=1\columnwidth]{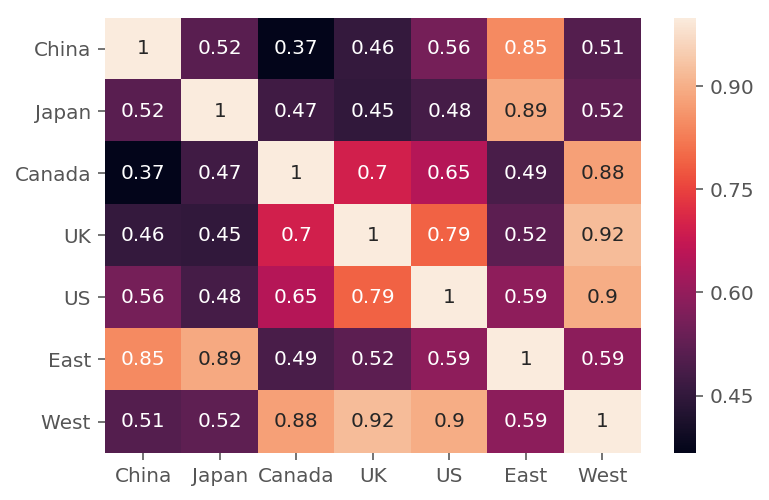}
\caption{Pairwise similarities (measured by Pearson $r$)
of countries in terms of Emojis learned from \textit{Word2Vec} models. East-West $r$ of $.59$ indicates some level of normativity, though lower than previous findings across Western languages.}
\label{fig:psm} 
\end{figure}

Vector representations allow for mathematical projections, which essentially
serve as a measure of similarity. We compute a pairwise similarity
for each pair of Emojis in each country, and use the vectors of per-country
pairwise Emoji similarities as the basis of generating a country-level
pairwise similarity matrix (shown in Figure \ref{fig:psm}). The
Pearson correlation coefficient between the West and the East is $0.59$,
indicating similarity in the semantics of Emoji usage even across
two different cultures. While this supports some level of normativity,
we find that this level of East-West similarity is lower than previous
findings~\cite{barbieri2016cosmopolitan} of Emoji semantics across
four Western languages where similarity matrices of Emojis were correlated
$>.70$. Our within Western nation correlations were similar to past
findings, ranging from $.65$ to $.79$. Altogether, these results
reveal that there is still normativity in Emoji usage across East-West
with the moderate positive Pearson correlation, though there is less
similarity than if we were to compare across Western nations.


\subsection{Association with Psycholinguistic Categories}

\begin{figure*}[t!]
	\center
	\includegraphics[width=.7\textwidth]{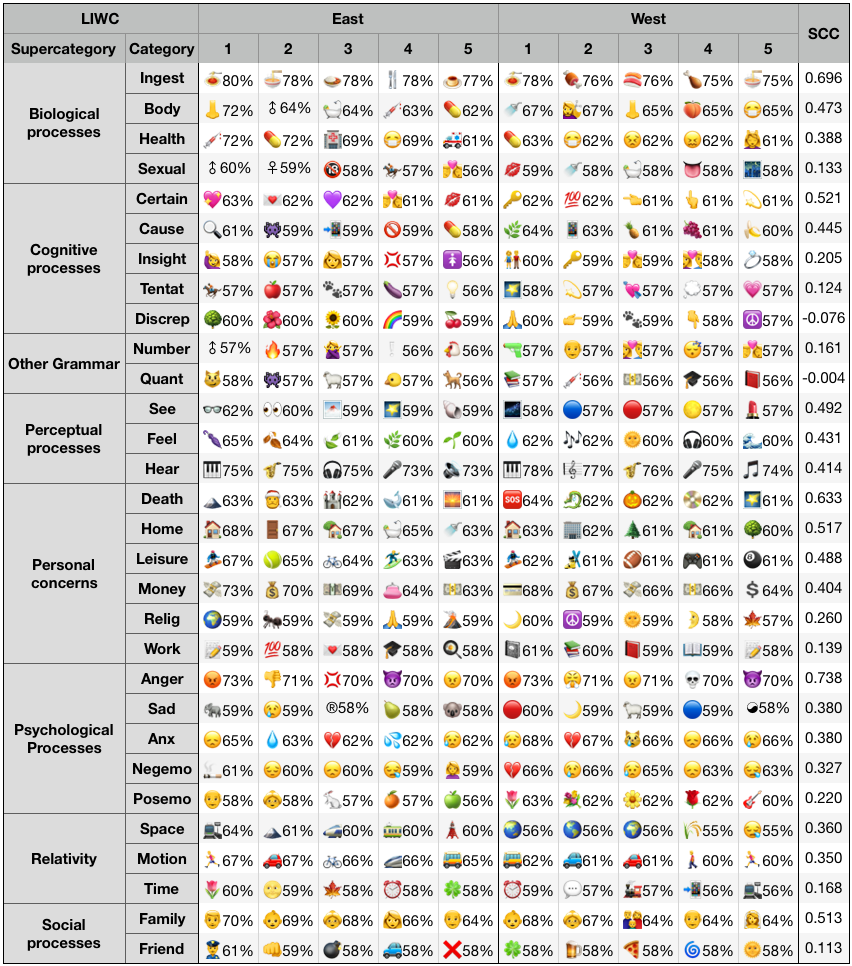}
	\caption{Association with Psycholinguistic categories represented by LIWC with top 5 Emojis in the Eastern countries and in the West, ranked by their similarity with each LIWC category. The SCC for all Emojis in each LIWC category is also presented on the right indicating a measure of corresponding (dis-)similarity. In each cell, the SCC is computed for the shown Emoji between the two cultures. The two arrays (on which this SCC is calculated) contain the cosine similarities of the Emoji vector and each LIWC category vector.}
	\label{fig:LIWC_all} 
\end{figure*}

Figure \ref{fig:LIWC_all} demonstrates top 5 Emojis similar to each
LIWC category in the East and in the West. Specifically, these results
reveal the correspondence between the LIWC category (and all its related
words) and a set of Emojis. The extent that SCCs are high shows that
the same set of words across two cultures relate to the same types
of Emojis; low SCCs reveal that the same category of words is associated
with different Emojis across the two cultures.

There is overall evidence for normativeness between East and West in how concepts captured by LIWC are represented by Emojis. Almost all the LIWC categories have positive SCCs and the median SCC is $.38$. At the same time, there are also specific categories that reveal more distinctiveness between the two cultures. In the next paragraphs, we describe specific findings in an exploratory manner.

\begin{figure*}[t!]
	\centering{}\includegraphics[width=0.7\textwidth]{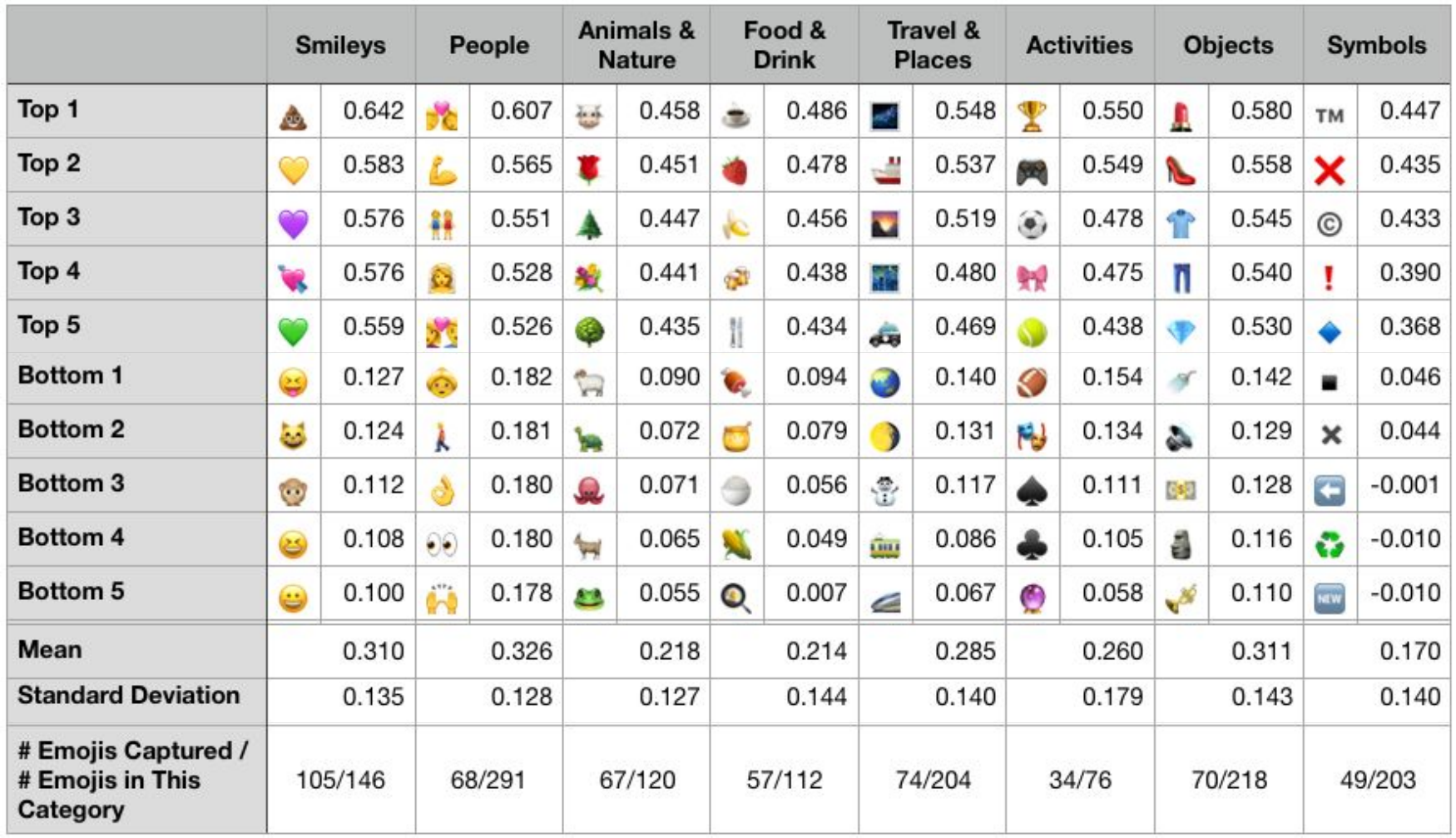}
	\caption{Top and Bottom 5 Most Universal Emojis in Terms of Similarities to
		LIWC Categories, grouped by Unicode Category. Emoji icon differences
		measured by Spearman correlation across East and West. Correlation
		is computed on similarities of the Emoji vectors to each of the LIWC
		category vectors in both corpora. Top 10 and bottom 10 correlated
		Emojis across both platforms are shown for each category in the Unicode
		Consortium, along with the mean and variance of each category.}
	\label{fig:icon_diff} 
\end{figure*}

\begin{figure*}[t!]
\centering{}\includegraphics[width=.9\textwidth]{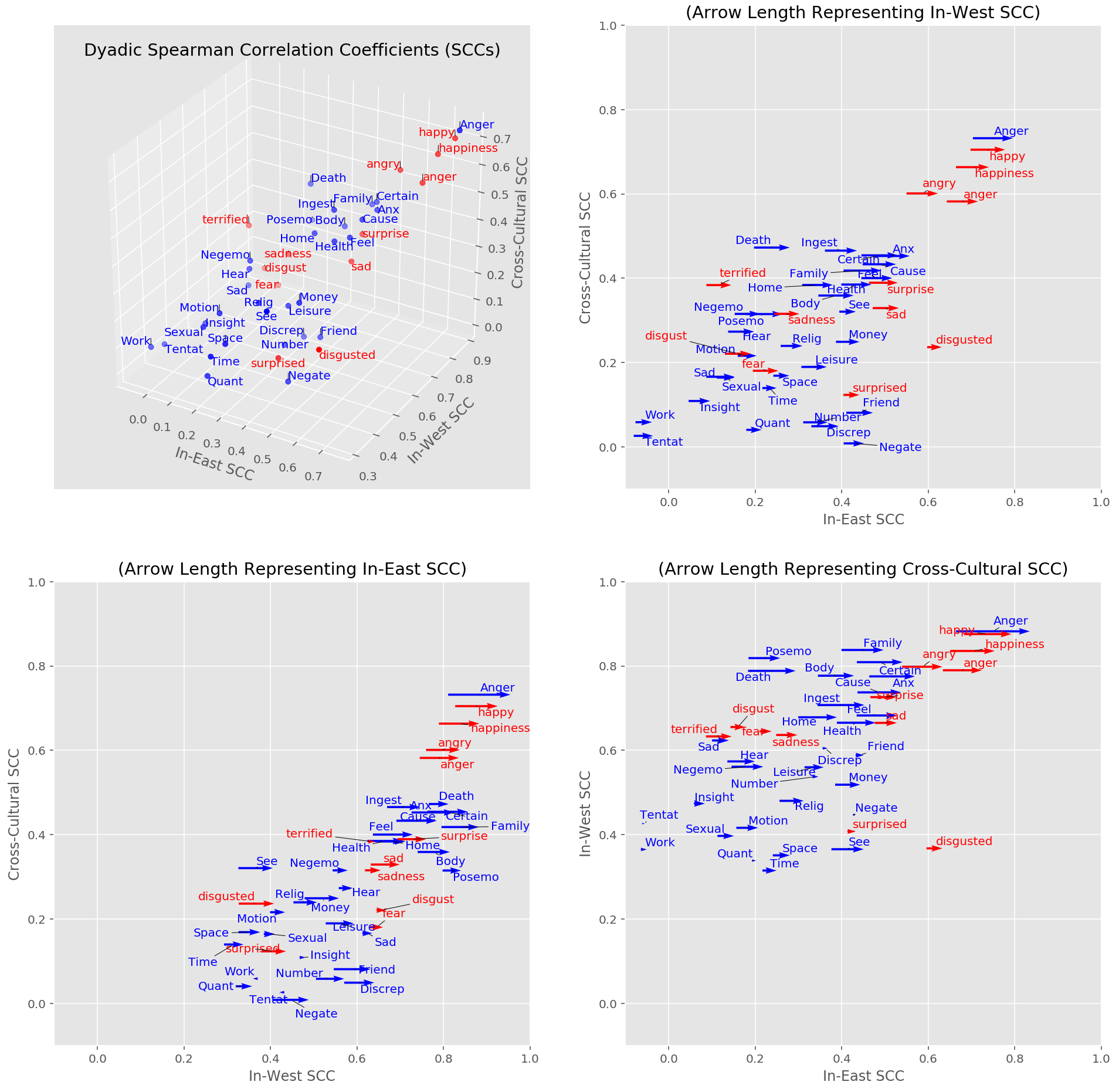}
\caption{SCCs of LIWC categories and Ekman emotion words using similarities with all Emojis as underlying values. The three axis of the scatter plot represent SCCs between the three western countries, SCCs between the the eastern countries, and SCCs between the West and the East, respectively. Both adjective and noun forms of the Ekman emotion words are considered and labeled in red. Points with blue labels are LIWC categories. Multi-view projection is shown as quiver plots, where arrows are colored under the same rules.}
\label{fig:quiver} 
\end{figure*}

Substantively, LIWC categories can be represented from words into Emoji expressions; the rank-order correlations reveal if these Emoji expressions overlap in reflecting the specific category. 

\paragraph{East-West Similarities.} LIWC categories that are most similar in terms Emojis are `Ingest', `Death', `Anger', `Money', `Home', and `Family'. \textit{Prima facie}, many of these categories are recognized as universal and the choice of Emojis to represent these categories are the most similar. Given that money is a medium of exchange in almost all societies of the world given global capitalism~\cite{berger1996national}, `Money' is a category that is universally understood and regarded
in a similar way and this is represented as such with
Emojis (💸, 💰, etc.). This also applies to the Emoji expressions in
the category of `Death' (🏰 and 🎅 in the East; 🆘, 🎃, etc. in the West) which is the ultimate issue all humans face~\cite{kubler1973death}.
Similarly, the categories of emotion `Anger' (😡, etc.) is tied to
the basic emotions of anger which has been found to be universally
expressed and recognized facially~\cite{Ekman1992argument}. The
category of Ingest (🍝, etc.) and how people imbibe food as expressed
in Emojis are also similar based on the rank-order correlations. 

\paragraph{East-West Differences.} At the same time, amid similarity overall, we also observe that there
are some cultural dimensions that emerge from the plots as the Emojis
for rice bowl (🍚) and ramen (🍜) dominated the East, while meat-related
Emojis (🍖, 🍗, etc.) take the majority in the West~\cite{prescott1995cross,ahn2011flavor}.
On the other hand, the LIWC categories that have lower correlations,
and indeed even, inverse correlations show that Emojis used to express
these constructs are likely different overall.
`Insight', `Discrepancy', `Quantitative', `Number', `Time', `Friend',
and `Work' have small or near-zero correlations. This seems to be
in line with categories that are linked to cultural influence. In
terms of the grammatical categories of `Quantifiers' and `Number',
given that there are differences in the grammar and syntax of Chinese,
Japanese, and English, this difference is also understandable. Similarly,
`Time' is often viewed symbolically and is laden with cultural meaning;
moreover, there are differences in the importance of keeping time
and the timing of events~\cite{brislin2003cultural}. East Asians
and Westerner also have differences in interpersonal dealings. With
regard to the former, Confucianism places a premium on harmony and
proper relationships as the basis for Asian society whereas Westerners
often place greater importance in outcomes and direct communication~\cite{yum1988impact}.
This is revealed in differences in the category of `Friend' in how
Emojis are used to express that idea. To a larger extent, it also
extends to the broader category of social processes where we also
find a low correlation on Emojis expressing the concept of `Family'.
Emojis also seem to reflect governmental policies. For example, the
Chinese government had been banning game consoles till 2015, and --
in our dataset collected from 2014 -- the game controller Emoji `🎮'
that dominates West in `Leisure' is nowhere to be seen in the East.

With regard to the categories of Discrepancy and Work having lower
correlations, our explanations are at best speculative. The category of `Discrepancy' (containing the textual tokens `should', `could', `would') may be expressed differently with Emojis due to the deferential culture and higher power-distance in the Eastern context as opposed to the Western context~\cite{farh2007individual,schwartz1994beyond}.

\subsection{Icon Differences}

Given that Emojis are based on an ideographic system where each symbol
represents a specific concept, it is also important to examine how
Emojis are similar or different between the Eastern and Western contexts.
For this, we transpose the results from association with psycholinguistic categories to determine
the extent Emojis are similar based on how the different LIWC categories
are projected onto the Emojis. Figure \ref{fig:icon_diff} shows the
top 5 and bottom 5 Emojis (ranked by SCC) across both platforms
for each Unicode category. The mean and std. dev of each of these
categories are also presented. Social scientific theories emphasizing
on the universality of basic emotions suggests that emotion-expressing
Emojis tend to show high convergence even between distinct cultures
such as the East and the West ~\cite{Ekman1992argument,Ekman2016scientists}.
Therefore, it is expected that categories of `Smileys' and `People'
would likely be more convergent compared to other categories. Consistent
with this expectation, the Emoji categories of `Smileys' and `People'
display relatively higher mean correlations between both cultures
($\rho=.31$ and $.32$ respectively).

A significant cultural difference component is language itself as
it forms the basis of cultural expression. The use of \textit{emics}, from within the social group, in anthropology and psychology where cultural behaviors and ideas are understood from the context of the culture itself emphasizes the
specificity and distinctiveness of language rather than its commonality~\cite{harris1976history}.
From this view, the `Symbols' category is likely to converge less.
This was borne out from the relatively low average correlations ($.17$)
between Eastern and Western cultures for the `Symbols' category.

It is again important to note that there appears to be evidence for
the universality of Emojis from this analysis as there is a positive
correlation across all the different Emoji categories. Further categories such as `Objects' and `Travel' had similar levels of correlations as `Smileys' and `People'.

\subsection{Association with Universal Ekman Emotions}

We further investigated the semantics of Emoji usage and how they
vary when compared against expression of universal basic
emotions (specifically Ekman categories). If Emoji
representations are normative, we would find similar levels of SCCs
with basic emotion categories. 
As LIWC does not cover all 6 Ekman basic emotions~\cite{Ekman1992argument}, we looked at specific emotion
words such as `anger' and `happy'. To overcome the selection bias
in part of speech, we considered both nouns and adjective forms. We
consider $12$ individual word vectors learned from each \textit{Word2Vec} model. Hence, we extend the previous definition of `category vector'
$\vec{c}_{l,i}$ to include also $12$ Ekman emotion words. For each
LIWC category and Ekman emotion word $i$, SCCs between country pairs
$\left\{ l_{1},l_{2}\right\} $ are computed. Represented by $\vec{s_{d}}_{l_{1},l_{2},i}$,
they are then averaged with respect to whether $l_{1}$ and $l_{2}$
are both from the East (`In-East SCCs'), both from the West (`In-West
SCCs'), or different cultures (`Cross-Cultural SCCs'). The 3 vectors
are plotted as coordinates in the 3D scatter graph in Figure \ref{fig:quiver}.
We find that, based on SCC magnitudes, there is a greater similarity
in Emoji representation among Western nations compared to the East.
LIWC categories such as `Friend', `Insight', `Motion', `Work', and
`Number' had relatively low similarities within Western and Eastern
contexts and also did not have substantial East-West similarity. However,
categories such as `Anger', `posemo' (positive emotion), `Death',
and `Family' had relatively higher similarities within Western and
Eastern contexts and also higher similarities across the two cultures.
Ekman emotion word terms were also included in the quiver plots to
assess the degree to which basic emotions are similar within and also
across Eastern and Western cultures. We found that the most universal
terms were with regard to anger and happiness (i.e., similarity within
and between cultures). However, with regard to surprise, disgust,
sadness, and fear there was less \textit{relative} similarity across
cultures. We emphasize \textit{relative} because this also confirms our findings
that the Emoji representations instantiated in LIWC categories have
a substantial degree of normativeness; therefore, we find that even
basic emotion categories (e.g., surprise, disgust, sad/sadness, fear/terrified)
do not uniquely distinguish themselves to have much higher cross-cultural
SCCs, although we do see a trend that some categories like `Quant',
`Time', `Work', and `Space' have lower cross-cultural convergence.

\subsection{Limitations and Future Work}

In this paper, we find, amid similarity, that there are some cultural
dimensions that emerge from how the semantics of Emoji vary across
both cultures. While these differences were studied primarily from
the perspective of Emoji use, a large portion of it could potentially
also be attributed to the text of the post. By training \textit{Word2Vec} models with both text and Emoji tokens across the East and the West,
and by analyzing the Emoji associations with word categories as captured
by LIWC, we attempt to uncover the interactions between both. However,
it would be interesting to study the cross-cultural variation in text
and Emoji usage independently to quantify each. Approaches from recent
work on understanding Emoji ambiguity in English~\cite{miller2017understanding} could be coupled with ours to achieve this goal.

Even though we attempted to compare Emoji usage in 2 Eastern countries
(Japan and China) and 3 Western countries (US, UK, and Canada) respectively, we nevertheless used two different platforms, namely Weibo and Twitter to represent each. We also used LIWC from Chinese and English, and obtained Japanese version by translating the Chinese LIWC. This could potentially introduce confounds around platform differences, over and above cultural differences. To minimize this, we restricted posts based on geo-location, dropped bilingual posts, and analyzed posts
in the primary language of communication in each country. Prior studies
also found a lot of similarities in user demographics, intention of
use and topical differences~\cite{gao2012comparative,ma2013electronic,lin2016exploring}. However, it would be promising to look at data from other sources (such as smart phone users~\cite{lu2016learning}) where social desirability and censorship confounds might be lower to further validate the findings in our study.

All our analyses were based on correlation rather than causality.
Because of the richness and diversity of the Emojis, it is difficult
to hypothesize \textit{a priori} how specific Emojis may differ. Therefore, we undertook an abductive approach to construct explanatory theories as patterns emerged from our analysis~\cite{haig2005abductive}. For social science research, these methods offer data-driven insights
into group and user behaviors which can be used to generate new hypotheses for testing and can be used to unobtrusively measure large populations over time. Commercial applications include improving targeted online marketing, increasing acceptance of Human Computer Interaction systems and personalized cross-cultural recommendations for communication.

Future work should investigate similarities and differences in other socio-psychological constructs. Also, considering the promise of Emojis in downstream application tasks such as sentiment analysis, studies should explore the contribution of Emojis in multi-modal and cross-lingual sentiment analysis and transfer learning tasks.

\section{Conclusion}

In this paper, we compared Emoji usage based on frequency, context,
and topic associations across countries in the East (China and Japan)
and the West (United States, United Kingdom, and Canada). Our results
offer insight into cultural similarities and differences at several
levels. In general, we found evidence for the normativeness, or the
universality, of Emojis. While there are relative differences in that
Western users tend to use more Emojis than Eastern users, the relative
frequencies in different types of Emojis are correlated across cultures.
Moreover, distributional semantics found that the Emoji expressions
were clustered in a similar manner across cultures. Even when we used
universal basic emotions as a benchmark, we found that Emojis were
represented in a cross-culturally similar manner compared to these
basic emotion expressions.

At the same time, we found that there appear to also interpretable
distinctions between Emoji use based on topical analyses. Emojis were
culturally specific as certain types of Emojis such as rice-based
dishes had the highest projection on the LIWC category of `Ingest'
in the East while a mix of meat and spaghetti had the highest project
on the same category in the West. Analysis at the icon level reveal
support for general social scientific theories of cultural similarities
and differences where relative similarities were found more in terms
of the `Smileys' and `People' icons whereas relative differences were
found for `Symbols' icons. Nevertheless, these findings need to be
construed from the perspective that there appears to be a robust thread
of cross-cultural similarity in Emoji patterns.

\bibliographystyle{aaai}
\bibliography{FULL-GuntukuS.1010}
\end{document}